\documentclass{article}
\usepackage[dvipsnames]{xcolor} 
\usepackage[preprint]{corl_2026} 
\usepackage{hyperref}
\usepackage{graphicx}
\usepackage{amsmath}
\usepackage{amssymb}
\usepackage{wrapfig}

\usepackage{tikz,pgfplots}
\pgfplotsset{compat=1.18} 
\usetikzlibrary{external} 
\usetikzlibrary{calc}

\usepackage{algorithm}
\usepackage{algorithmic}

\usepackage{multirow}
\usepackage{booktabs}
\usepackage{subcaption}

\usepackage{placeins}

\usepackage{pifont}   
\newcommand{\cmark}{\ding{51}}
\newcommand{\xmark}{\ding{55}}

\DeclareMathOperator*{\argmin}{arg\,min}

\title{VIGOR-MPC: Vision-Guided Objective \\ Regulation MPC}

\author{
  Jelena Trisovic\textsuperscript{*,1,2}\\
  \texttt{tjelena@ethz.ch} \\
  \And
  Rahel Rickenbach\textsuperscript{*,1} \\
  \texttt{rrahel@ethz.ch} \\
  \AND
  Marek Landert\textsuperscript{1} \\
  \texttt{marek.landert@bluewin.ch} \\
  \And
  Panagiotis Minos\textsuperscript{1} \\
  \texttt{minospanagiotis@gmail.com} \\
  \And
  Melanie N. Zeilinger\textsuperscript{1} \\
  \texttt{mzeilinger@ethz.ch} \\
}

\begin{document}
\maketitle

\renewcommand{\thefootnote}{\arabic{footnote}}
\footnotetext[1]{
Institute for Dynamic Systems and Control, ETH Zurich, Switzerland, \textsuperscript{2} ETH AI Center, \textsuperscript{*} These authors contributed equally (shared first authorship).\textsuperscript{3}The codebase is available in the eth research collection repository, \url{https://doi.org/10.3929/ethz-c-000807583}, or under \url{https://github.com/VIGOR-MPC/codebase/}. Further material can be found under \url{https://vigor-mpc.github.io/}.}


\begin{abstract}
Controlling an agent in an environment solely perceived through onboard vision is challenging: perception yields noisy, variable-length, and partially occluded context, while onboard compute is shared between perception and control, limiting the planning horizon executable in real time. We propose VIGOR-MPC, a modular student–teacher framework, agnostic to the perception pipeline, which {addresses reference tracking across navigation tasks by distilling} a privileged, field-of-view-aware long-horizon MPC into a computationally lightweight short-horizon MPC whose cost parameters are determined online from vision. The teacher operates in simulation with full state knowledge and a multi-trajectory formulation that guarantees the agent can stay within its field of view. The student observes only the noisy, occluded context produced by a perception module and learns, via a {multi-level} recurrent mapping, to infer cost parameters that aim to recover the teacher's behavior despite missing information. We evaluate VIGOR-MPC on vision-based autonomous racing in simulation and on hardware. Using only onboard perception of the track, VIGOR-MPC generalizes to tracks unseen during training and approaches the lap times of the privileged long-horizon teacher while reducing per-step computation by over  90\%.\textsuperscript{3} 

\end{abstract}



\section{Introduction}

In vision-based control tasks, an agent acts in an unknown environment that it perceives only through an onboard camera. While this setting allows for operation without previously mapping the environment and thereby enables a broader range of applications, such as hand-eye coordination for robotic grasping from monocular images \citep{levine2018learning}, or general exploration tasks of mobile robots \citep{sridhar2024nomad}, it introduces challenges for control performance and safety: (i) uncertain and noisy perception, leading to imperfect reference and constraint estimates, (ii) a limited and potentially occluded field of view, (iii) the computational overhead of the perception module leaving limited compute for the controller, which can hinder fast reaction times, and (iv) the absence of prior knowledge about the environments encountered at deployment. While challenges (i)–(iii) are inherent to any vision-based controller, challenge (iv) is where end-to-end learning approaches fundamentally struggle, as they tend to generalize poorly to unseen layouts or new information sources.\\
Inspired by recent progress in combining vision-based with optimization-based control~\citep{loquercio2021learning,trisovic2025uncertainty}, as well as developments in imitation learning exploiting the structure of optimization-based controllers~\citep{amos2018differentiable,rickenbach2025zipmpc, song2022policy}, we demonstrate how structured imitation learning can be leveraged to bring the strengths of optimization-based control to the vision-based setting.  
Our approach employs a student-teacher framework that addresses these challenges through a privileged, field-of-view-aware model predictive control (MPC) teacher with full state information, that is by construction capable of remaining safely within the field of view, and a computationally efficient student {with multi-level memory} that, via imitation, learns to adapt the parameters of a predefined control structure from the noisy, partially occluded, variable-length context produced by a perception module. Unlike prior work that learns short-horizon cost parameters under perfect information~\citep{rickenbach2025zipmpc}, our setting introduces imperfect, time-varying, {partially occluded and missing} context, addressed through a field-of-view-aware privileged teacher, a memory-capable parameter mapping, and real-time context-driven reference adjustment mechanisms enabling obstacle avoidance from perceived context alone. The student thereby inherits the teacher's field-of-view-aware safe and obstacle avoidant behavior while adapting online to perceptual uncertainty, demonstrated in simulation and on hardware on a vision-based racing task with no prior track knowledge.

\paragraph{Related Work}
We organize prior work on vision-based control by how much classical control structure is preserved between perception and action. \textit{End-to-end policies} map camera observations directly to actions, from early behavior-cloning approaches in driving~\citep{bojarski2016end,codevilla2018end} to recent foundation-model-scale variants~\citep{hu2023gaia}. However, they are data-intensive, lack constraint-handling, and require retraining under domain shift. \textit{Modular pipelines with a classical controller} feed fixed-form perception outputs into a predefined controller, which was applied both in agile flight~\citep{kaufmann2018deep,loquercio2019deep,loquercio2021learning}
and driving~\citep{mueller2018modularity,chen2019lbc}. The controller is safe and interpretable but rigid, and cannot adapt to noisy, or variable-length context. A recent line of work designs controllers that handle perception uncertainty explicitly~\citep{trisovic2025uncertainty,modares2025perception,dromc2025edl} through different variants of constraint tightening rather than learned
adaptation. \textit{Modular pipelines with a learned controller} either map
perception-derived system and environment states to actions via reinforcement learning or imitation trained policy~\citep{kaufmann2023champion,wurman2022outracing,lee2020quadruped,xing2024bootstrapping}
or learn only specific ingredients of an otherwise classical
controller, such as the cost or decision variables of an
MPC~\citep{song2022policy,romero2024actor,rickenbach2025zipmpc},
dynamics
model~\citep{nagabandi2018neural}, its sampling
distribution~\citep{sacks2023learning}, or a terminal value
function~\citep{lowrey2018plan}. Fully learned
controllers handle constraints only implicitly; partially learned ones
preserve the optimization scaffolding and inherit its safety and
generalization, but have been developed almost exclusively with clean
state-based context, hence neglecting perception. Building on~\citep{rickenbach2025zipmpc},
our work fills the gap in this last category by adapting cost
parameters to the noisy, occluded, variable-length context produced
by a perception module.

\begin{figure}
    \centering
    \includegraphics[width=0.99\linewidth]{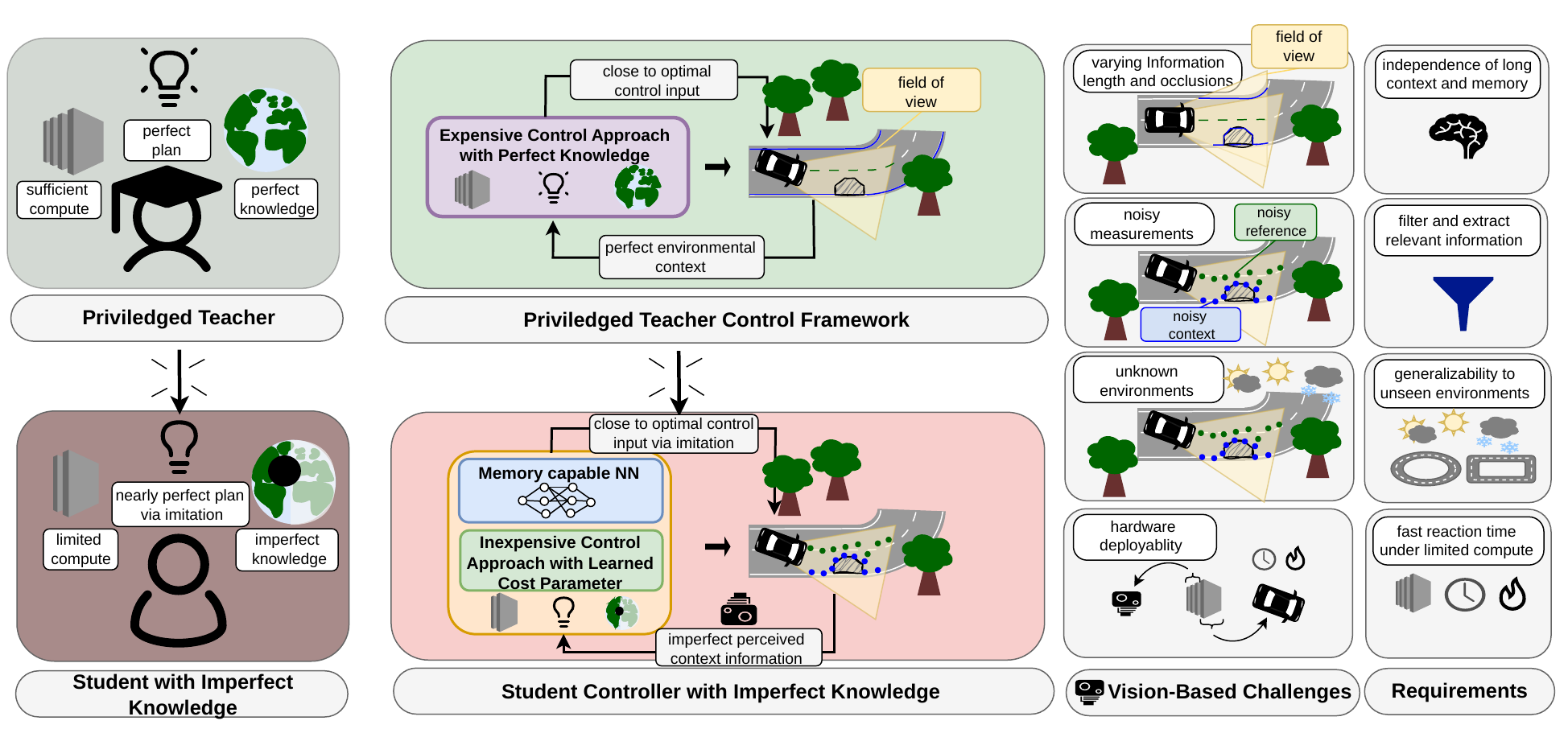}
    \caption{Overview of the VIGOR-MPC framework. The student employs a predefined, parameterized MPC structure whose cost parameters are generated by a memory-capable neural network from noisy and partial environmental observations. The mapping is trained via imitation learning using a privileged teacher with full environmental knowledge. The vision-based challenges addressed by the framework, together with the resulting control requirements, are illustrated on the right.}
    \vspace{-0.7cm}
    \label{fig:priv_mpc_explanation}
\end{figure}

\section{Problem Formulation}
\label{sec:requirements}

We consider an agent with nonlinear dynamics $x(k+1) = f(x(k),u(k),\zeta_d(k))$, where $x \in \mathbb{R}^n$, $u \in \mathbb{R}^m$, and $\zeta_d  \in \mathbb{R}^{a}$ denote the state, input, and the dynamics-relevant
environmental context at time $k \in \mathbb{N}$, respectively. Other environmental information that has no direct impact on the dynamics, but potentially affects the constraints, such as obstacles, is denoted by $\zeta_e(k)$. The objective is to ensure safe motion while tracking a reference trajectory $\mathbf{r}_{N_k}(k) = [r(k), r(k+1), \hdots, r(k+N_k)]$ over the remaining task length $N_k$,   given corresponding, analogously denoted context sequences $\boldsymbol{\zeta}_{d,N_k}(k)$ and $\boldsymbol{\zeta}_{e,N_k}(k)$. {The agent's
state $x(k)$ is assumed known, while reference and contextual information are extracted from the partial, noisy observations of an onboard RGB-D camera, resulting in four challenges and corresponding requirements for the controller, as illustrated in Figure~\ref{fig:priv_mpc_explanation}.}

\textbf{R1 - Safety under Limited Context with Variable Length and Occlusions:} When a camera is mounted on a moving agent, information about upcoming references and the environment is limited by its field of view. Furthermore, as the agent, camera, or surrounding objects move, the field of view changes and previously visible information may disappear. This occurs, for example, in autonomous driving when a vehicle enters a curve, in tabletop manipulation when a robotic arm occludes objects and reference points, {or when perceived frames are lost during transmission}. Instead of the required $\textbf{r}_{N_k}(k)$, $\boldsymbol{\zeta}_{d,N_k}(k)$, and $\boldsymbol{\zeta}_{e,N_k}(k)$ we rather observe $\textbf{r}_{\tilde{N}^r_k}(k)$, $\boldsymbol{\zeta}_{d,\tilde{N}^d_k}(k)$, and $\boldsymbol{\zeta}_{e,\tilde{N}^e_k}(k)$, with $\tilde{N}^r_k$, $\tilde{N}^d_k$, and $\tilde{N}^e_k$ smaller or equal to $N_k$, potentially different from each other, {and sometimes even equal to zero}. Moreover, as the agent moves through the unknown environment, the amount of captured contextual information varies with its orientation, and in general $\tilde{N}^i_k \neq \tilde{N}^i_{k+1}$ for $i \in \{r,d,e\}$. Consequently, the controller must (i) ensure safe operation by planning the existence of a trajectory within the field of view and enforcing constraint satisfaction under a time-varying effective horizon $\tilde{N}^i_k \leq N_k$, $i \in \{r,d,e\}$, without conditioning on a fixed or minimum lookahead; (ii) maintain a recurrent internal state that aggregates past observations, so that context which is temporarily occluded or out of view at time $k$ remains available to the controller.

\textbf{R2 - Robustness to Noisy and Uncertain Context:} Relevant environmental information must be extracted from the captured images using a suitable perception algorithm, which introduces
a source of uncertainty that manifests as noise and outliers in the estimated references and context. Such perception errors are typically difficult to model or bound reliably, as they combine small estimation noise with occasional large outliers and may vary with viewing angle, lighting, or scene content.  
The resulting estimates are denoted as $\mathbf{\hat{r}}_{\tilde{N}^r_k}(k) = \mathbf{r}_{\tilde{N}^r_k}(k) + \mathbf{\epsilon_{r}}$, $\boldsymbol{\hat{\zeta}}_{d,\tilde{N}^d_k}(k) = \boldsymbol{\zeta}_{d,\tilde{N}^d_k}(k) + \epsilon_{\zeta_{d}}$, and $\boldsymbol{\hat{\zeta}}_{e,\tilde{N}^e_k}(k) = \boldsymbol{\zeta}_{e,\tilde{N}^e_k}(k) + \epsilon_{\zeta_{e}}$, where $\mathbf{\epsilon_{r}}$, $\epsilon_{\zeta_{d}}$ and $\epsilon_{\zeta_{e}}$ indicate the noise vectors of corresponding dimensions  whose distributions are generally unknown
and non-stationary. The controller should therefore compensate for perception uncertainty directly from data, without requiring explicit uncertainty estimates, prescribed error bounds, or distributional assumptions on $\epsilon$.

\textbf{R3 - Generalizability to Unknown Environments and Modularity:} The
core motivation for vision-based control is that the environment is unknown both during deployment and training, and hence the framework cannot be trained on the exact target environment. Furthermore, the representations of $\hat{\mathbf{r}}$, $\boldsymbol{\hat{\zeta}}_d$, and $\boldsymbol{\hat{\zeta}}_e$ are application-dependent: the context may encode any handcrafted or learned property of the scene obtained from arbitrary sensor modalities, while the reference may be specified, e.g., as waypoints or road curvature. The controller should hence generalize to previously unseen environments and remain agnostic to the specific perception pipeline producing its inputs.

\textbf{R4 - Fast Reactions under Limited Compute:} In practical hardware deployments, computational resources are inherently limited and control decisions must be generated with low latency. {At the same time, the perception stack already imposes a substantial computational burden, leaving only limited resources for the control computation. The resulting controller should hence be computationally efficient, ensuring fast reaction times while maintaining high performance. }

Since racing applications on an unknown track \citep{cai2021vision, trisovic2025uncertainty} describe a safety-critical robotics application in an unknown, continuously changing environment, they describe a representative benchmark for the proposed framework. Hence, for illustration, we will repeatedly refer to the example of an autonomous car, operating on an unknown planar road of constant and known width~$2\omega$. \begin{wrapfigure}{r}{0.28\textwidth}
\vspace{-0.3cm}
    \centering
    \includegraphics[width=0.30\textwidth]{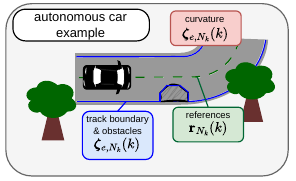}
    \caption{Illustration of the considered example setup.}
    \label{fig:exampleintro}
    \vspace{-0.5cm}
\end{wrapfigure} In this setting, $\mathbf{r}_{N_k}(k)$ is the track centerline and $\boldsymbol{\zeta}_{e,N_k}(k)$ contains the track boundaries and any road-blocking obstacles.  The vehicle dynamics are expressed in the Frenet frame attached to the centerline, so that the local track curvature enters $f(\cdot)$ explicitly. Hence, $\boldsymbol{\zeta}_{d,N_k}(k)$ is the track curvature along the upcoming reference. The example setup is visualized in Figure~\ref{fig:exampleintro} and further details are provided in the Appendix~\ref{apx:implementation_details}. {To demonstrate that VIGOR-MPC is not specific to car racing, we additionally include quadrotor racing and reference-tracking experiments in Appendix~\ref{apx:drone_results}.}

\section{Method}
\label{sec:method}
The proposed framework adopts a student–teacher imitation learning structure, specifically designed to address the challenges inherent to vision-based control introduced in Section~\ref{sec:requirements}. Since both the student and the teacher build upon MPC, we briefly introduce the MPC problem in Section~\ref{subsec:preliminaries}. This is followed by a detailed overview of the proposed learning framework and the employed key concepts in Section~\ref{subsec:frameworkoverview}. A visualization of the main idea is provided in Figure~\ref{fig:priv_mpc_explanation}. 

\subsection{Preliminaries: Model Predictive Control}
\label{subsec:preliminaries}
With $N$ indicating the prediction horizon, we denote a general MPC controller with $\mathit{MPC}_{N}({C_N}, \textbf{r}_{N}(k), \boldsymbol{\zeta}_{N}(k), x(k))$. Combining both context sources in $\boldsymbol{\zeta}_{N}(k) = [\boldsymbol{\zeta}_{d,N}(k),\boldsymbol{\zeta}_{e,N}(k)]$, at time step $k$, it takes the current state $x(k)$, as well as the reference and context information $\textbf{r}_{N}(k)$, $\boldsymbol{\zeta}_{d,N}(k)$, and $\boldsymbol{\zeta}_{e,N}(k)$ as input and solves:
\begin{equation}
\label{eq:mpclh}
\begin{aligned}
 \relax  [X_{k,N}^*, U_{k,N}^*] &:= \argmin_{\{x_i\}_{0}^{N+1}, \{u_i\}_{0}^{N}} \sum_{i=0}^{N} \ell(x_i, u_i,c_i,r_i)  \\
    \text{s.t.} \quad x_{0} = x(k), \quad x_{i+1} &= f(x_i,u_i,\zeta_{d,i}), \quad x_i \in \mathcal{X}(\zeta_{e,i}), \quad u_i \in \mathcal{U},  \quad i=0,\hdots,N.
\end{aligned}
\end{equation}
The function $\ell$ defines the stage cost 
with parameters $c_i \in \mathbb{R}^{b}$, summarized in $C_{N} := \{c_i\}_{0}^{N}$.
Furthermore, potentially context-dependent state and control-input constraints are indicated with $\mathcal{X}(\zeta_{e,i}) \subseteq \mathbb{R}^n$ and $\mathcal{U}\subseteq \mathbb{R}^m$, respectively. Solving Eq.~\ref{eq:mpclh} yields the optimal state and control-input trajectories over the prediction horizon $N$, denoted by $X_{k,N}^* := \{x_i^*\}_{0}^{N+1}$ and $U_{k,N}^* := \{u_i^*\}_{0}^{N}$, with $x_i^* \in \mathcal{X}(\zeta_{e,i})$ and $u_i^* \in \mathcal{U}$ indicating the optimal state and control input at prediction step $i$, respectively. Then, only the first optimal control input $u_0^*$ is applied, after which the optimization is reinitialized and solved again at the next time step.

\subsection{Imitation Learning from Perfect Knowledge}
\label{subsec:frameworkoverview}
The proposed approach leverages a privileged teacher operating in a near-perfect setting, while the student adheres to a predefined control structure and learns to adapt its objective parameters. This enables the student to imitate the teacher’s full knowledge behavior under imperfect, real-world conditions and partial knowledge. Since the proposed approach is designed to work in an unknown environment, we train in a representative example environment and leverage the student's generalization capabilities to transfer to unseen ones. We thereby design the components as follows:  

\textbf{Environment:} The training setting should be representative of the expected target environment, but does not have to fully align with it. It should share the same types of references and contextual information, while ideally covering a broad range of the context variations expected during deployment. Moreover, to enable both teacher supervision and student training, it must provide access to both perfect ground-truth information and a realistic, noisy perception of the scene. We thus simulate the investigated robotics platform and perception pipeline, providing ground-truth references and context $\mathbf{r}_{N_k}(k)$, $\boldsymbol{\zeta}_{d,N_k}(k)$, $\boldsymbol{\zeta}_{e,N_k}(k)$ at any time step $k$, as well as their noisy and potentially occluded counterparts $\mathbf{\hat{r}}_{\tilde{N}^r_k}(k)$, $\boldsymbol{\hat{\zeta}}_{d,\tilde{N}^d_k}(k)$, $\boldsymbol{\hat{\zeta}}_{e,\tilde{N}^e_k}(k)$ obtained from the simulated perception pipeline.

\textbf{Teacher:} The teacher is privileged: it has access to sufficient computational resources,  and operates on ground-truth references and context $\mathbf{r}_{N_k}(k)$, $\boldsymbol{\zeta}_{d,N_k}(k)$, $\boldsymbol{\zeta}_{e,N_k}(k)$ at any time step $k$. Nevertheless, it remains aware of the limited field of view and employs a control approach that guarantees safety under all circumstances, satisfying the constraints and ensuring to remain within the current field of view. Under these assumptions, we design the teacher as a multi-trajectory model predictive controller (MT-MPC) \citep{saccani2022multitrajectory}. At every time step $k$, the MT-MPC guarantees the existence of a feasible control sequence that keeps the agent safely within the observable field of view. This is achieved by jointly predicting two trajectories, a safe trajectory $[X_{k,s,N}^*, U_{k,s,N}^*]$ and a performance trajectory $[X_{k,p,N}^*, U_{k,p,N}^*]$, coupled only through their first prediction step via the constraint $x_{p,1} = x_{s,1}$. The safe trajectory is constrained to remain within the field of view through a terminal constraint $x_{s,N_L+1} \in \mathcal{F}_{T}$, while the performance trajectory optimizes the control objective. We choose a long horizon $N_L$ and denote the resulting controller as $\mathit{MPC}_{MT,N_L}(C_{N_L}, \mathbf{r}_{N_L}(k), \boldsymbol{\zeta}_{N_L}(k), x(k))$. The full optimization problem is provided in Appendix~\ref{subsec:apx_mtmpc}.

\begin{figure}[htb!]
    \centering
    \includegraphics[width=0.99\linewidth]{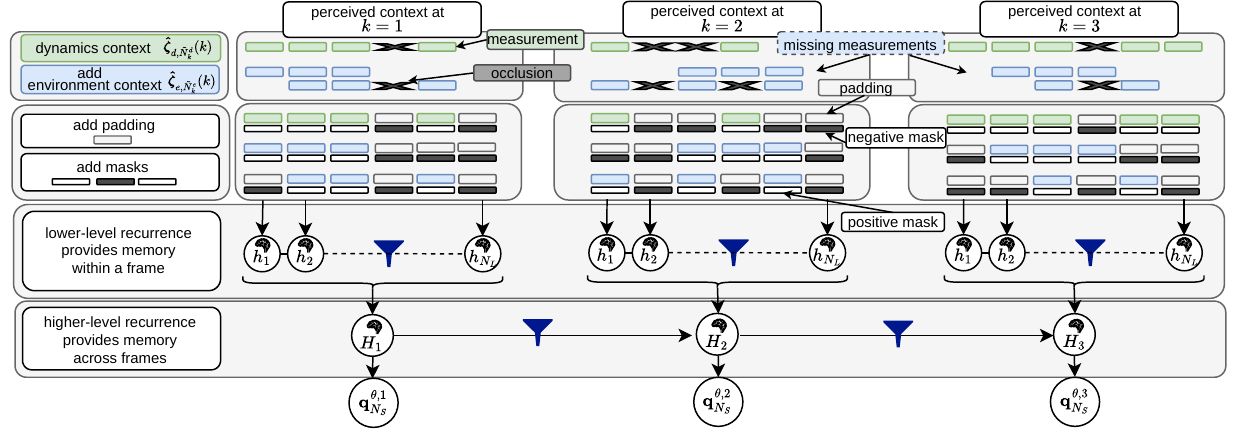}
    \vspace{-0.5em}
    \caption{Schematic overview of the proposed context-processing pipeline. To obtain a fixed-dimensional representation, missing entries are padded with non-informative values and accompanied by binary masks indicating valid observations. The resulting context sequence is processed by an RNN, which extracts relevant temporal information while maintaining an internal memory state, enabling retention of previously observed but currently occluded context.}
    \vspace{-0.5em}\label{fig:rnn_explanation}
\end{figure}

\textbf{Student:} The student operates in a simulation of the real-world (or in the real-world for finetuning) and receives noisy, partially occluded reference and context information $\mathbf{\hat{r}}_{\tilde{N}^r_k}(k)$, $\boldsymbol{\hat{\zeta}}_{d,\tilde{N}^d_k}(k)$, $\boldsymbol{\hat{\zeta}}_{e,\tilde{N}^e_k}(k)$ within the limited field of view. It is deployed as a closed-loop controller with a predefined, parametrized control structure whose parameters $\mathbf{q}_{N_S}^{\theta,k}=\{q_{i}^{\theta, k}\}_{i=0}^{N_S}$ are produced by a parametric mapping $\mathcal{R}_{\theta}$ as $\mathbf{q}_{N_S}^{\theta, k} = \mathcal{R}_{\theta}(x(k),\boldsymbol{\zeta}(k))$ from the noisy context and the current state, resulting in a controller denoted by  $\mathit{MPC}_{N_S}\!\left(\mathbf{q}_{N_S}^{\theta, k},\, \mathbf{\hat{r}}_{\tilde{N}^r_k}(k),\, \boldsymbol{\hat{\zeta}}_{d,\tilde{N}^d_k}(k),\, \boldsymbol{\hat{\zeta}}_{e,\tilde{N}^e_k}(k),\, x(k)\right).$ This construction follows the structural template of~\citep{rickenbach2025zipmpc}, in which a short-horizon MPC equipped with a learned, context-dependent cost imitates the behavior of a long-horizon MPC at a fraction of the computational cost. This template already addresses three of the requirements identified in Section~\ref{sec:requirements}: the MPC backbone provides the structural affordances for safe behavior,  such as constraint- and dynamics-aware predictions, partially addressing R1; the modular separation between the learned mapping and the MPC yields a controller that generalizes across environments and remains agnostic to the perception pipeline producing its inputs~(R3); and the short-horizon design leaves real-time budget for that perception pipeline~(R4). The field-of-view-aware safety behavior itself, however, is not inherent to this structure and must be instilled through imitation of the privileged teacher. 
Moreover, the original formulation assumes perfect, noise-free, and fully observable context, and therefore does not capture the limited and time-varying field of view or the perceptual uncertainty inherent to vision-based control. In addition, obstacle avoidance is not explicitly considered. We address these limitations along three axes, illustrated schematically in Figure~\ref{fig:rnn_explanation}. First, to enable reactive obstacle avoidance, we augment the environmental context with $\boldsymbol{\zeta}_{e,N_S}$ and allow the parametric mapping to influence the controller either directly, through constraint adaptation, or indirectly, through reference modification. Second, to represent partially observable context sequences within a fixed-dimensional input space, occluded measurements are padded using a non-informative value, while the corresponding visibility information is encoded through a binary mask for each context dimension.
Third, to retain information that was previously observed but has since left the field of view, we implement $\mathcal{R}_{\theta}$ as a recurrent neural network (RNN) {with multi-level recurrence given by parametrized nonlinear maps $\rho_{\theta}^{\mathrm{low}}$ and $\rho_{\theta}^{\mathrm{high}}$. The lower-level recurrence processes the spatial context within each perception frame $k$, with hidden states evolving as $h_i^k = \rho_{\theta}^{\mathrm{low}}(\zeta_i^k, h_{i-1}^k)$, where $\zeta_i^k$ is the input at context step $i$. Attention pooling aggregates these hidden states into a fixed-dimensional frame representation $s_k$, which is concatenated with the state $x(k)$, forming a higher-level input $e_k$. The higher-level hidden state then evolves as $H_k = \rho_{\theta}^{\mathrm{high}}(e_k, H_{k-1})$, and the network output is $\mathbf{q}_{N}^{\theta,k} = g_{\theta}(H_k, x(k))$. The lower-level recurrence thus provides memory within a frame and the higher-level one across frames, allowing the network to exploit observations that temporarily leave the field of view and providing a nonlinear filtering mechanism under partial observability and occlusions.}

\textbf{Training:} {We seek parameters $\theta^*$ of $\mathcal{R}_\theta$ that enable the student to imitate the privileged teacher despite noisy and limited observations, i.e., to learn to filter, distrust, and infer unknown parts of its context. To account for the memory of the multi-level recurrent architecture, training samples consist of sequences of $T$ consecutive frames. For a sequence ending at frame $k$, $\mathcal{R}_\theta$ processes the preceding $T-1$ frames to build its hidden state before predicting the cost parameters $\mathbf{q}_{N_S}^{\theta,k}$ for the final frame, on which the loss is evaluated. Specifically, we minimize the mean-squared error $\mathcal{L}$ between the predicted trajectories of the long-horizon MT-MPC teacher with full environmental knowledge, $[X_{k,p,N_L}^*, U_{k,p,N_L}^*] := \mathit{MPC}_{MT,N_L}(C_{N_L}, \mathbf{r}_{N_L}(k), \boldsymbol{\zeta}_{N_L}(k), x(k))$, and those of the short-horizon student MPC with the inferred cost parameters, $[X_{k,N_S}^{\theta,*}, U_{k,N_S}^{\theta,*}] := \mathit{MPC}_{N_S}(\mathbf{q}_{N_S}^{\theta,k}, \hat{\mathbf{r}}_{\tilde{N}^r_k}(k), \hat{\boldsymbol{\zeta}}_{d,\tilde{N}^d_k}(k), \hat{\boldsymbol{\zeta}}_{e,\tilde{N}^e_k}(k), x(k))$:
\begin{equation}
\label{eq:problem}
    \theta^* := \argmin_{\theta} \mathbb{E}_{k} \left[\mathcal{L}\!\left([X_{k,N_S}^{\theta,*}, U_{k,N_S}^{\theta,*}],\, [X_{k,p,N_L}^*, U_{k,p,N_L}^*]\right)\right],
\end{equation}
where the expectation is taken over initial states $x(k)$. We draw sequence starting points uniformly at random and retain only those for which the teacher admits a feasible solution over the full sequence. Since teacher and student horizons differ, the loss is evaluated over the first $N_D \leq N_S$ steps only. To backpropagate through the student MPC, we employ the differentiable-MPC approach of \cite{amos2018differentiable}, as in \cite{rickenbach2025zipmpc}. The full training algorithm is given in Appendix~\ref{apx:learningframeworksetup}.}

\section{Experiments}

We evaluate the proposed framework on a vision-based autonomous racing task used as an example throughout the paper. The system dynamics $f$ are modeled using a \emph{kinematic bicycle} model \cite{rajamani2011vehicle}, with the vehicle state and track expressed in the Frenet frame \cite{milliken1995race}. The controller is formulated as a model predictive contouring controller (MPCC) \cite{liniger2015optimization} with a quadratic stage cost $\ell$ and cost parameter vector $c_i = [q_i, p_i]$, where $q_i, p_i \in \mathbb{R}^{n+m}$. The context variable comprises a long-horizon sequence of track curvatures and track boundaries, potentially tightened by obstacles, within the field of view. Implementation details on the vehicle model and control formulation are provided in Appendix~\ref{apx:implementation_details}. \\
\textbf{Experiment Design and Baselines:} We evaluate our approach in simulation and on hardware, and compare against three baselines: (i) the privileged teacher $\mathit{MPC}_{MT,N_L}$ with $N_L=25$, i.e., a long-horizon MT-MPC with perfect environmental knowledge (reference and constraints) 
, (ii) a long-horizon MT-MPC with imperfect knowledge due to inferring it from images, $\mathit{MPC}_{MT^*,N_L}$, and (iii) a short-horizon MPC with $N_S=6$, without context-dependent adjustment of the cost parameters $\mathbf{q}_{N_S}^{\theta, k}$, denoted as $\mathit{MPC}_{N_S}$. Design choices for all baselines are detailed in Appendix~\ref{apx:implementation_details}{, as well as comparisons with a fully learned baseline.} Simulation results are presented in Section~\ref{sec:sim_results}, and hardware experiments in Section~\ref{sec:hardware}. Furthermore, additional simulation and hardware results, including the application of VIGOR-MPC to quadrotor dynamics, can be found in Appendix~\ref{apx:additional_results}.

\subsection{Simulation Results}
\label{sec:sim_results}
The racing track images are rendered in Blender~\citep{blender}, from which the perception module (Appendix~\ref{apx:perception}) extracts noisy reference and context information for the student. The following results show how the proposed framework meets the requirements in Section~\ref{sec:requirements}, i.e., robustness to imperfect knowledge, computation time reduction, behavior under limited and varying information length, and generalization. {Additionally, to motivate our design choices, we provide an ablation study of the method components.}

\begin {figure}[t]
\centering
\begin{tikzpicture}[scale=0.9]

    \node (pic1) at (-2.4, 0.0) {\includegraphics[width=.255\textwidth]{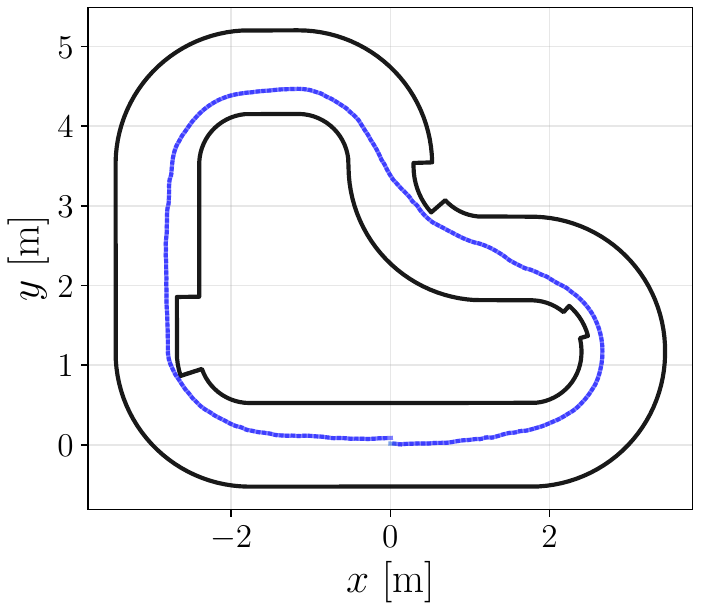}};

    \node (pic2) at (2.1, 0.0) {\includegraphics[width=.23\textwidth]{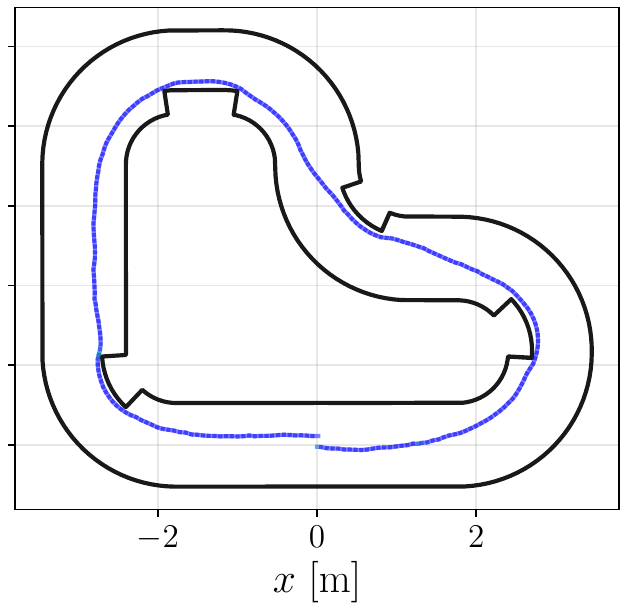}};

    \node (pic2) at (6.8, 0.0) {\includegraphics[width=.28\textwidth]{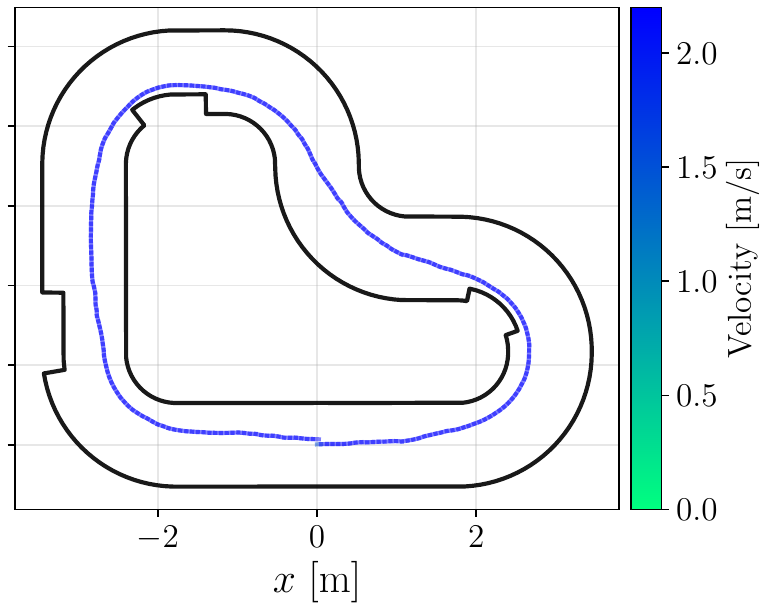}};

\end{tikzpicture}
    \caption{Visualization of environments with randomly generated obstacle configurations and closed loop application of VIGOR-MPC.} 
\label{fig:obstacle_avoidance}
\end{figure}

\textbf{Matching Teacher Performance under Imperfect Knowledge:} We evaluate each baseline against the privileged teacher $\mathit{MPC}_{MT,N_L}$ and provide the results  in Table~\ref{tbl:results_im_time_comp}.  The track layout is chosen as in Figure~\ref{fig:obstacle_avoidance}, without obstacles. We observe that VIGOR-MPC reduces the imitation error by $57.53\%$ over the short-horizon MPC and closely matches the imperfect-knowledge MT-MPC denoted with $\mathit{MPC}_{MT^*,N_L}$, despite operating on the same noisy context with a far shorter horizon. Additionally, VIGOR-MPC nearly matches the lap time of $\mathit{MPC}_{MT,N_L}$, while the short-horizon baseline is $7.5\%$ slower, and achieves a $94.2\%$ reduction in computation time, confirming that it nearly recovers teacher performance at a fraction of the computational cost. {Due to the observed heavy-tailed nature of the imitation error distribution, we provide further statistics in Table~\ref{tab:additional_stats} of the Appendix~\ref{subsec:add_sim_results}, reflecting reduced median and upper-tail error of VIGOR-MPC compared to $\mathit{MPC}_{N_S}$.}
\begin{table}[t]
\vspace{-1.5em}
\centering
\small
\caption{Imitation error computed as a root mean square error (RMSE) on states and control inputs over the first $5$ prediction steps with respect to $\mathit{MPC}_{MT,N_L}$, lap time, and per-step execution time, both reported as absolute values and relative changes with respect to the teacher $\mathit{MPC}_{MT,N_L}$. Mean and standard deviation over $50$ initial states for the lap time and $200$ sample states for the imitation error and execution time calculation. Evaluations are done on the track of Figure~\ref{fig:obstacle_avoidance} without obstacles.}
\renewcommand{\arraystretch}{1.15}
\setlength{\tabcolsep}{6pt}
\begin{tabular}{l||c|cc|cc}
\toprule
Controller & Imitation error $\downarrow$ & \multicolumn{2}{c|}{Lap time [s] $\downarrow$} & \multicolumn{2}{c}{Execution time [ms] $\downarrow$} \\
\midrule
$\mathit{MPC}_{MT^*,N_L}$  & $\mathbf{0.0103 \pm 0.0216}$           & $\mathbf{7.39 \pm 0.02}$ & $+0.4\%$ & $ 545.0 \pm 250.8$\textsuperscript{\textdagger} & ${0.0 \pm 46.0\%}$  \\
$\mathit{MPC}_{N_S}$       & $0.0219 \pm 0.0315$           & $7.91 \pm 0.05$ & $+7.5\%$ & $35.3 \pm 4.1$ & $93.5 \pm 1.1\%$ \\
VIGOR-MPC                    & ${0.0126 \pm 0.0335}$  & ${7.44 \pm 0.03}$ & ${+0.6\%}$ & $\mathbf{31.6 \pm 4.7}$ & $\mathbf{94.2 \pm 1.0\%}$ \\
\bottomrule
\multicolumn{6}{l}{\footnotesize\textsuperscript{\textdagger} Computed over $198$ initial states; $2$ outliers in which the MPC failed to converge were removed.}
\end{tabular}
\label{tbl:results_im_time_comp}
\vspace{-1em}
\end{table}

\textbf{Limited and Varying Information Length and Safety:} In Figure~\ref{fig:obstacle_avoidance} we illustrate environments with randomly generated obstacle configurations and show how VIGOR-MPC achieves reactive obstacle avoidance with respect to arbitrary and potentially changing obstacle placements. From the investigated 100 configurations, it successfully avoided all obstacles for 99 of them, and remained safe by coming to a complete stop rather than crashing on the remaining one. 

\begin {figure}[t]
\centering
\begin{tikzpicture}[scale=1.0]

    \node (pic1) at (-2.4, 0.0) {\includegraphics[width=.3\textwidth]{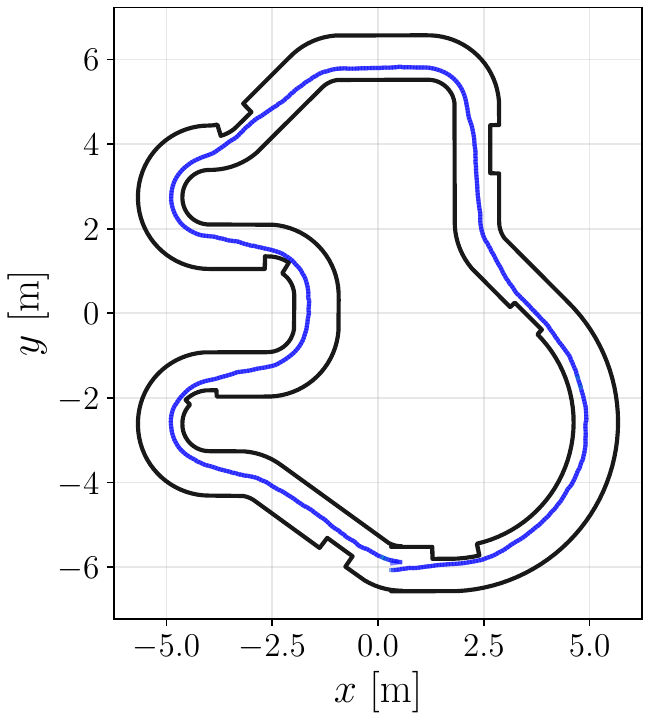}};

    \node (pic2) at (1.11, -1.05) {\includegraphics[width=.17\textwidth]{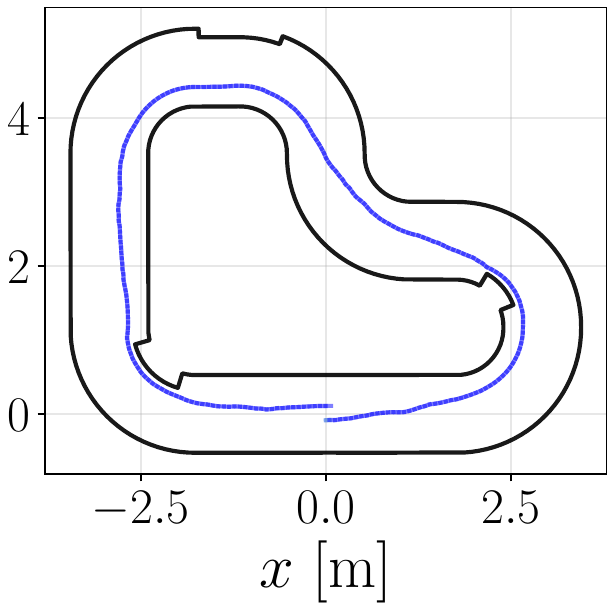}};

    \node (pic2) at (1.05, 1.2) {\includegraphics[width=.18\textwidth]{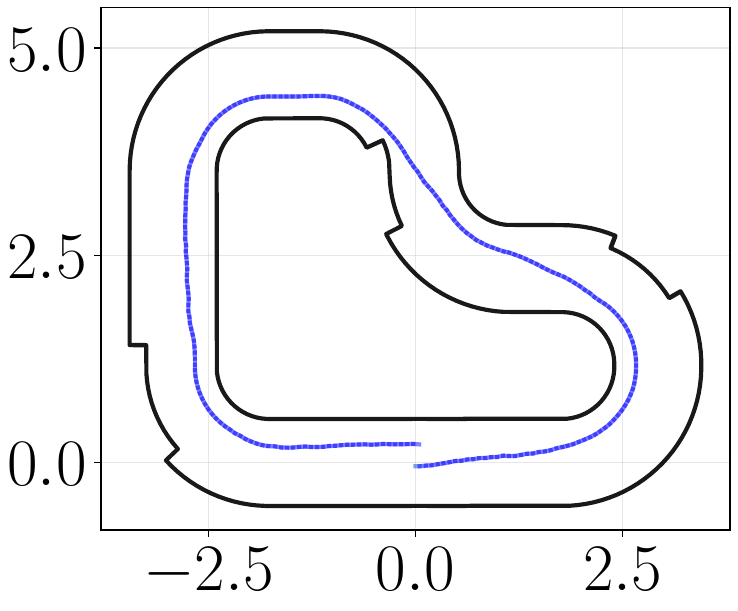}};

    \node (pic2) at (4, -0.67) {\includegraphics[width=.2\textwidth]{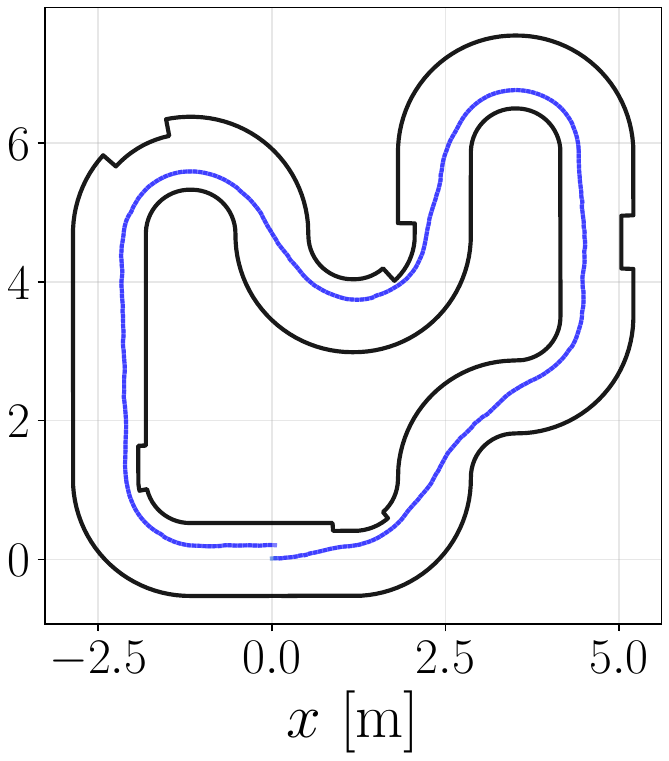}};

    \node (pic2) at (7, -0.67) {\includegraphics[width=.2\textwidth]{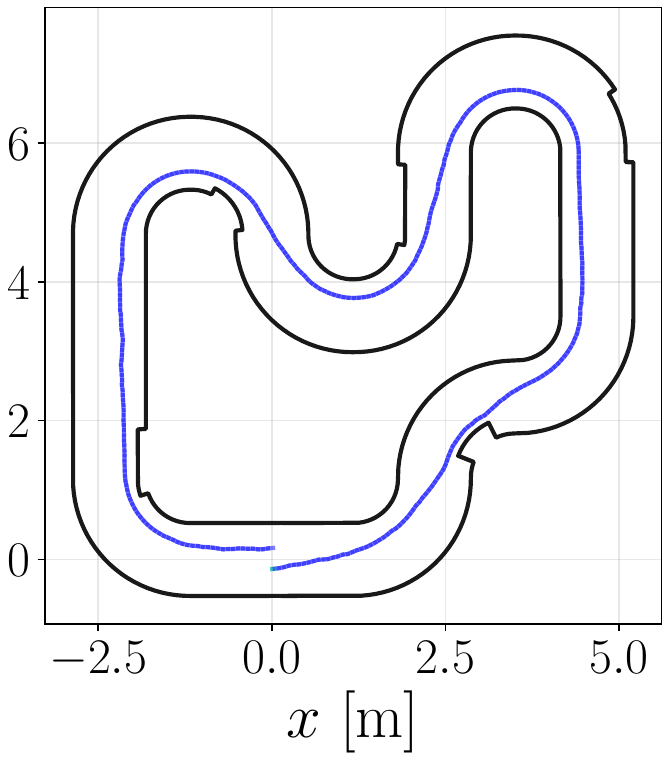}};

        \node (pic2) at (5.5, 1.6) {\includegraphics[width=.35\textwidth]{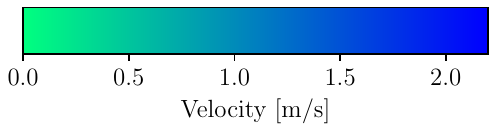}};

\end{tikzpicture}
    \caption{{Closed-loop employment of VIGOR-MPC on three different track layouts (with randomly placed obstacle configurations), where only the track layout on the left was seen during training.}} 
    \vspace{-1em}
\label{fig:generalization}
\end{figure}

\textbf{Generalization to Unseen Environments:} The generalization capabilities of VIGOR-MPC are visualized in Figure~\ref{fig:generalization}, showing the training and different test track layouts and obstacle configurations with successful, i.e., performant and obstacle-avoiding, closed-loop application of VIGOR-MPC. The corresponding  lap times are reported in Table~\ref{tbl:generalization}. 

\begin{table}[htb!]
\centering
\small
\caption{Lap time comparison of VIGOR-MPC and perfect teacher on two unseen tracks with two object configurations each.}
\renewcommand{\arraystretch}{1.15}
\begin{tabular}{l||c|cc|cc}
\toprule
 & Train & \multicolumn{2}{c|}{Test Track 1} & \multicolumn{2}{c}{Test Track 2} \\
Controller & Track & top & bottom & left & right \\
\midrule
$\mathit{MPC}_{MT,N_L}$    & ${17.75}$ s  & $7.36$ s& $7.46$ s& $11.29$ s& $11.29$ s\\
VIGOR-MPC                    & $18.12$ s & $7.39$ s&  $7.46$ s& $11.32$ s& $11.32$ s \\
\bottomrule
\end{tabular}
\label{tbl:generalization}
\end{table}

\begin{table}[t]
\centering
\small
\caption{{Ablation over the FOV-aware components, starting from an obstacle-aware ZipMPC baseline. Each row cumulatively enables the corresponding component, reporting lap completion rate over
100 random obstacle configurations.}}
\label{tab:ablation}
\begin{tabular}{cccc c}
\toprule
FOV-Aware MT-MPC Teacher & Masking & Context Recurr. & Time Recurr. & Success \\
\midrule
\xmark & \xmark & \xmark & \xmark & $0\%$  \\
\cmark & \xmark & \xmark & \xmark & $4\%$  \\
\cmark & \cmark & \xmark & \xmark & $3\%$  \\
\cmark & \xmark & \cmark & \xmark & $6\%$  \\
\cmark & \cmark & \cmark & \xmark & $46\%$ \\
\cmark & \cmark & \cmark & \cmark & $94\%$ \\
\bottomrule
\end{tabular}
\end{table}

\begin {figure}[H]
\centering
\begin{tikzpicture}[scale=0.9]

    \node (pic1) at (-3.2, 0.0) {\includegraphics[width=.25\textwidth]{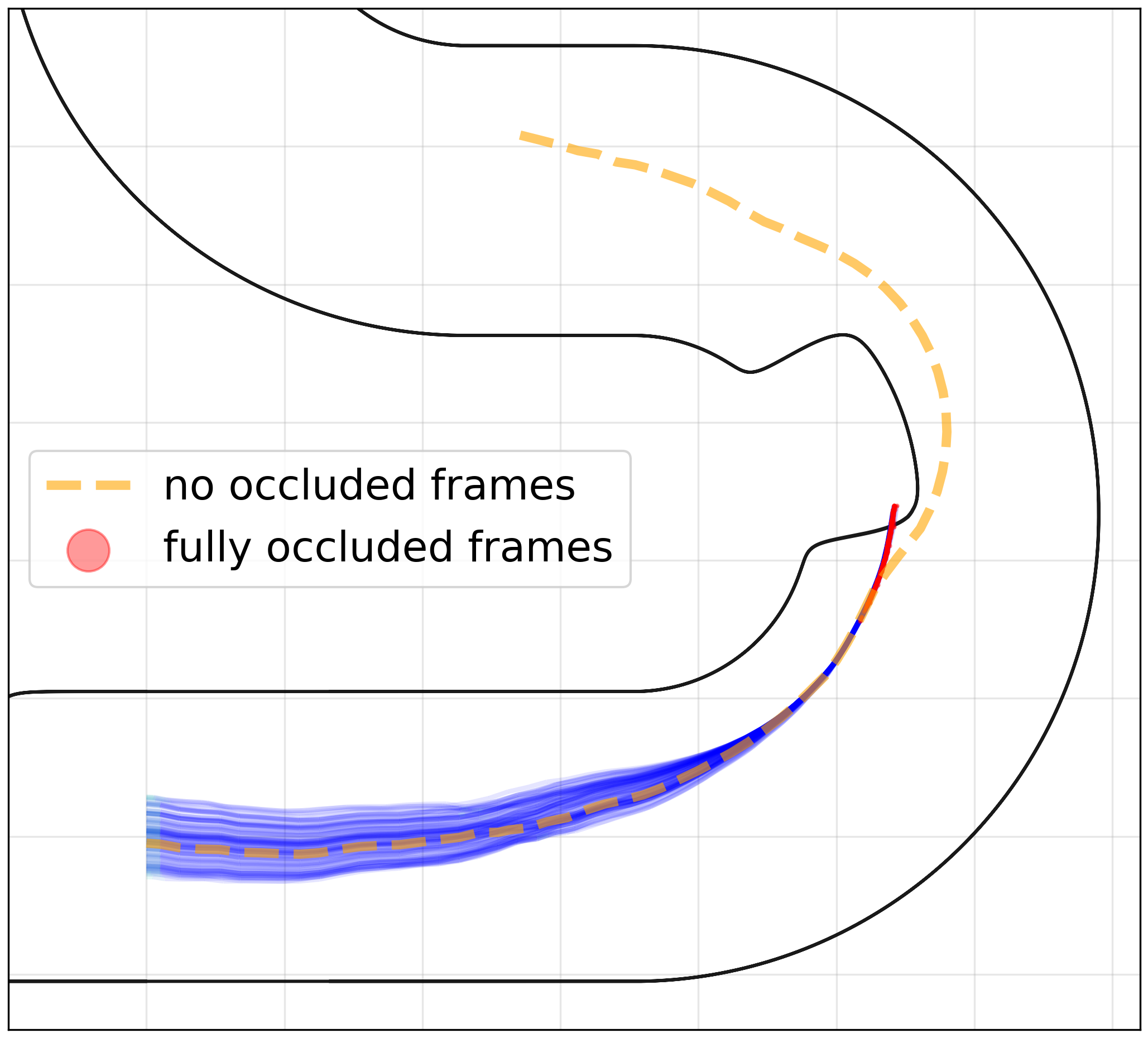}};

    \node (pic2) at (1.05, 0.0) {\includegraphics[width=.25\textwidth]{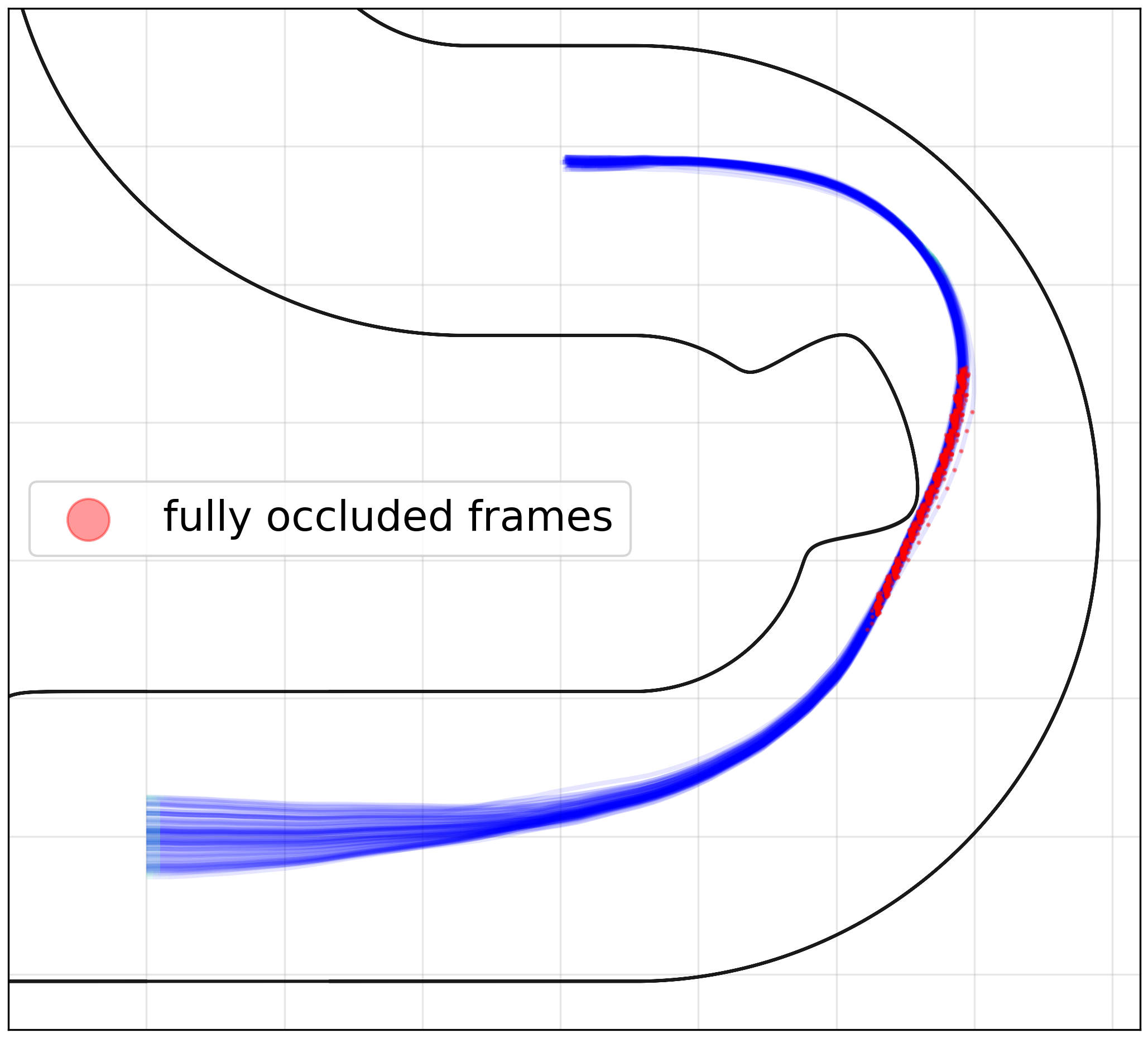}};   

        \node (pic2) at (6.04, 0.0) {\includegraphics[width=.335\textwidth]{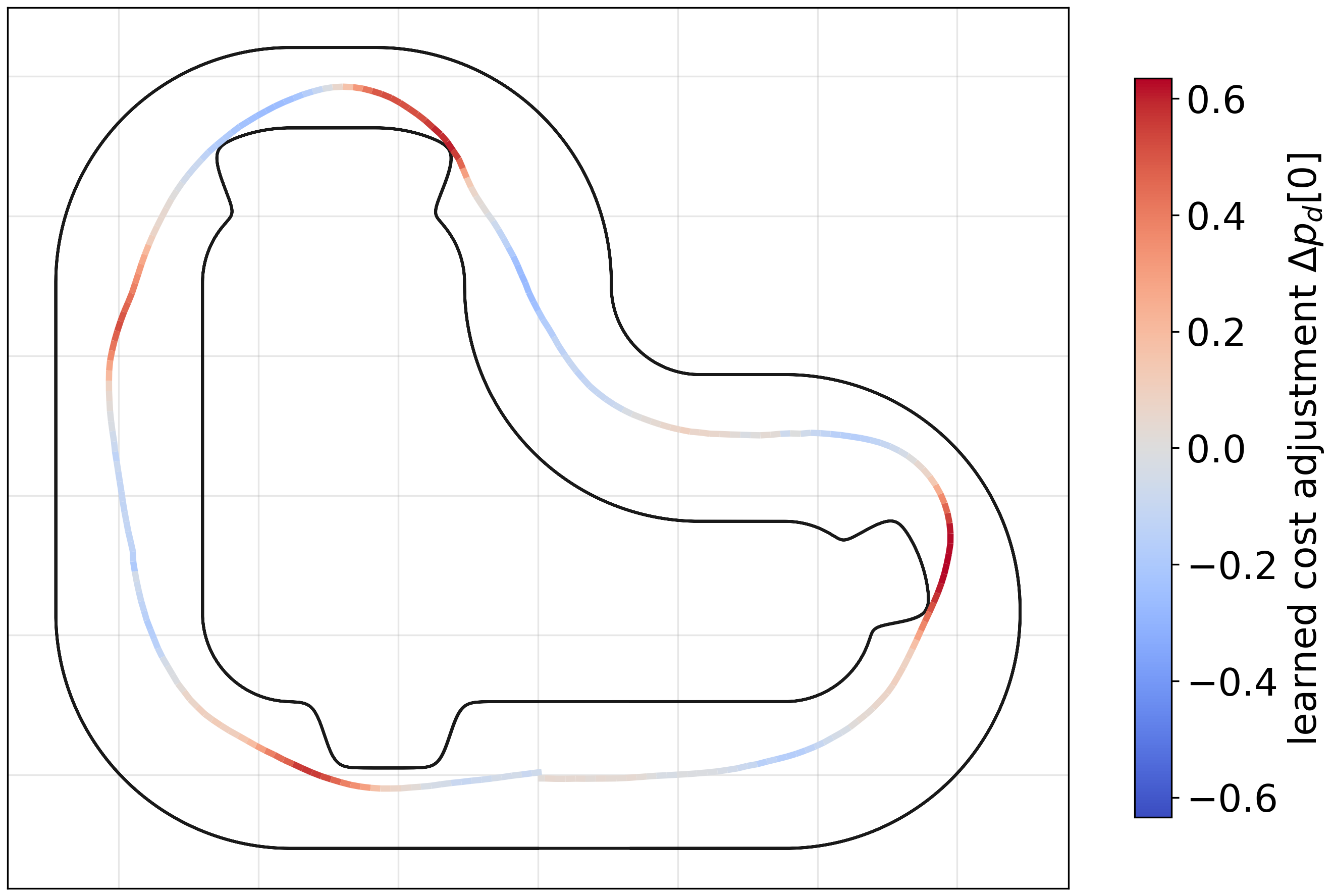}};   
\end{tikzpicture}
    \caption{{(Left and center) Impact of multi-frame recurrency for 100 initial conditions: critical frame drops right before an obstacle handled successfully when multi-frame memory is available. (Right) Lateral cost adjustment shifts references.}} 
    \vspace{-1em}
\label{fig:occlusion}
\end{figure}

\textbf{Ablation Study:} We ablate the components that distinguish VIGOR-MPC from its ZipMPC backbone~\citep{rickenbach2025zipmpc}. Starting from an obstacle-aware ZipMPC (with context padding), VIGOR-MPC introduces four elements: (a)~a field-of-view-aware MT-MPC teacher in place of ZipMPC's plain long-horizon teacher; (b)~binary masks marking which entries of the padded context are genuine observations rather than non-informative padding values; (c)~a context recurrence, i.e., the RNN unrolled over the context sequence within a single frame; and (d)~a time recurrence, i.e., the hidden state propagated across
frames, acting as a memory of context that has left the field of view. Table~\ref{tab:ablation} reports the lap completion rate
as these components are added cumulatively, over 100 random obstacle
configurations deliberately harder than those of
Figure~\ref{fig:obstacle_avoidance}: twice as many obstacles, wider on average, and predominantly on the inner side of the track.  From the baseline, adding the FOV-aware teacher alone raises completion only marginally, and adding either masking or context
recurrence individually on top of it yields little. The two are strongly complementary: combined, they reach $46\%$, as the mask tells the recurrence which observations to trust while the recurrence is what exploits that information. Finally, the cross-frame time recurrence achieves the success rate of $94\%$.

We further isolate the time recurrence under complete frame occlusions: in Figure~\ref{fig:occlusion} we drop the $10$ frames preceding an obstacle once it has entered the field of view, comparing VIGOR-MPC with and without the multi-frame recurrence. The same figure illustrates how  cost term $p_d$ that penalizes lateral deviation reshapes the centerline reference in response to newly perceived context.

\subsection{Hardware Experiments}
\label{sec:hardware}

  \begin{wrapfigure}{r}{0.28\textwidth}
\vspace{-0.5cm}
    \centering
    \includegraphics[width=0.3\textwidth]{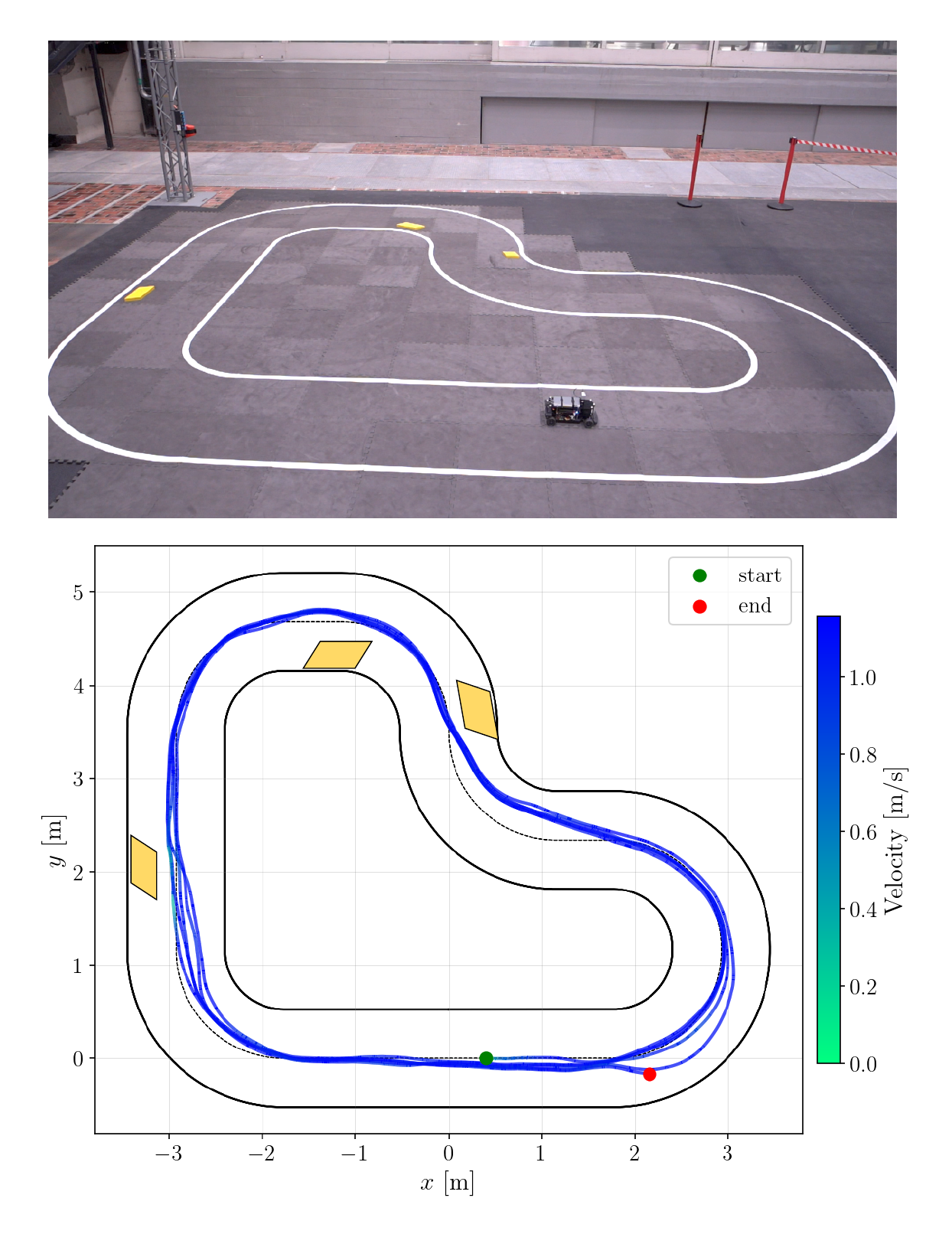}
    \caption{Closed-loop deployment of VIGOR-MPC on hardware.}
    \label{fig:hardware_with_img}
    \vspace{-0.5cm}
\end{wrapfigure}  For the real-world experiments, we employ a modified F1Tenth platform~\citep{o2020f1tenth} using the software stack of~\cite{carron2023chronos}. Onboard computation is performed on a LattePanda Sigma equipped with an Intel Core i5-1340P processor, ensuring that perception and control run locally without external compute. The MPC is solved with acados~\citep{verschueren2022acados} at a control rate of 30 Hz. From the Intel RealSense D405 RGB-D camera stream, a perception pipeline detects the lane markings and estimates the centerline, while obstacles are handled by a separate detection module; further details are provided in Appendix~\ref{apx:perception}. The track is built from interlocking foam mats with a lane width of 1.10 m, and lane boundaries are defined by white adhesive tape on a black surface to provide high contrast for perception. Ground-truth state estimates are obtained from an external motion capture system. A visualization of the obtained closed-loop performance of VIGOR-MPC is provided in Figure~\ref{fig:hardware_with_img}, including an image of our real-world testing setup at the top. The perception pipeline and all training data were generated entirely in simulation using Blender-rendered environments, with no real-world data used during training. Further details and results from our real-world experiment, including generalization capabilities, behavior under dynamic obstacle replacement, and comparisons with $MPC_{N_S}$ can be found in Appendix~\ref{apx:hardware_setup} and~\ref{apx:additional_results}. 

\section{Conclusion}
\label{sec:conclusion}

In this work, we present VIGOR-MPC, an imitation learning framework that combines a privileged teacher with a student leveraging the structure and advantages of optimization-based control with learned adaptation to address the challenges arising from vision-based control. The privileged teacher operates with perfect environmental knowledge and a long-horizon MT-MPC, while the student leverages a predefined short-horizon MPC structure whose cost parameters are generated by a memory-capable neural network from partial and noisy observations. 
The proposed framework was evaluated on the example application of autonomous racing. The results demonstrate that VIGOR-MPC can successfully approximate the behavior of the privileged teacher despite a limited field of view and imperfect perception, in simulation and the real-world.

\section{Limitations}
Despite the promising results, the proposed framework exhibits several limitations that motivate future research. First, the current setup relies on a motion capture system for velocity estimates. However, the car's relative position and orientation with respect to the track are estimated directly from onboard perception. In principle, velocity estimates could be obtained via point-cloud registration and an Extended Kalman Filter instead. Second, with the current context choice, obstacles are restricted to the left or right side of the track and cannot occupy the center line. More expressive context choices or richer MPC parameterizations could allow handling more complex obstacle configurations. Third, although the proposed imitation learning framework inherits desirable properties from the privileged teacher, strong theoretical guarantees cannot be provided without exact imitation across the full feasible state space.

\acknowledgments{This research was partly supported by the ETH AI Center through an ETH AI Center doctoral fellowship to Jelena Trisovic. Additionally, it was supported as a part of NCCR Automation, a National Centre of Competence in Research, funded by the Swiss National Science Foundation (grant number 51NF40\_225155).}

\bibliography{example}  
\newpage

\appendix

\section{Extended Related Work}
\label{sec:extended_related_work}

We position our approach within the broader landscape of vision-based
control by organizing prior work depending on {how much
classical control structure is preserved between perception and control}.

\paragraph{End-to-end policies.}
At one extreme, a single neural policy maps camera observations
directly to actions. Early work cloned human driving demonstrations
end-to-end~\citep{bojarski2016end,codevilla2018end}, and the paradigm
has since been scaled to video- and language-conditioned foundation
models trained on fleets of real-world driving
data~\citep{hu2023gaia}. While such policies can acquire complex
sensorimotor behaviors, training requires substantial data and
compute, the resulting policies offer no explicit mechanism for
constraint satisfaction, and non-trivial changes to the deployment
domain or task typically necessitate retraining. Robotics applications
that demand fast training, modest compute, and reuse of existing
perception modules therefore favor modular architectures that separate
perception from control. This modularity is empirically dominant across robotics, including agile flight, autonomous driving, and legged locomotion.

\paragraph{Modular pipelines with a classical controller.}
Classically, a modular architecture is adopted where the perception module produces a
fixed-form plan (waypoints, trajectories, or segmentations) that a
classical controller executes. Such pipelines have been deployed
across robotics, including drone
racing~\citep{kaufmann2018deep,loquercio2019deep}, agile flight in
cluttered environments~\citep{loquercio2021learning}, and autonomous driving~\citep{mueller2018modularity,chen2019lbc}, where these perception outputs are tracked by a fixed low-level controller. These systems
inherit the safety and interpretability of a non-learned controller
and transfer well to real hardware, but the controller is rigid, as it
cannot adapt to features of the perceived context that are noisy,
occluded, or of variable length. A growing line of work therefore
designs controllers that explicitly make use of perception
uncertainty, for instance by propagating sampled curvature
realizations through a multi-stage
MPC~\citep{trisovic2025uncertainty}, enclosing perception noise within
set-based observers for tube-based MPC~\citep{modares2025perception},
or integrating evidential confidence into a distributionally robust
MPC~\citep{dromc2025edl}. Such methods share our motivation of
treating perception output as a non-ideal signal, but address
uncertainty through constraint tightening rather than learned
adaptation of the controller itself.

\paragraph{Modular pipelines with a learned controller.}
Between these extremes lie modular pipelines in which the controller
is itself partly or fully learned. {Fully learned} controllers
map a perception-derived state estimate to actions via a neural network
trained with RL, imitation, or a combination of the two; prominent
examples include
Swift~\citep{kaufmann2023champion}, GT~Sophy~\citep{wurman2022outracing},
and teacher-student distillation pipelines for legged locomotion and
agile flight~\citep{lee2020quadruped,xing2024bootstrapping}. The
resulting controller, however, remains unstructured: constraints and
partial observability are handled only implicitly, and adaptation to
new deployment conditions still demands extensive data collection and
retraining. {Partially learned} controllers instead preserve an
optimization-based structure and learn only specific ingredients of
it, such as the cost or decision variables of an
MPC~\citep{song2022policy,romero2024actor,rickenbach2025zipmpc}, the
dynamics model~\citep{nagabandi2018neural}, the sampling
distribution~\citep{sacks2023learning}, or a terminal value
function~\citep{lowrey2018plan}.
Because the dynamics model, constraints, and optimization scaffolding
are retained, such controllers inherit the generalization,
interpretability, and constraint-handling properties of classical
control while still adapting to context from data.

Our work belongs to this last category. The methods above, however,
have been developed almost exclusively with clean, low-dimensional
context from ground-truth simulation or accurate state estimation. To
our knowledge, none has been deployed in a vision-based regime where
the context is itself the noisy output of a perception module. Our approach addresses this gap, building on~\citep{rickenbach2025zipmpc} and extending it with ingredients required for vision-based deployment, detailed in Section~\ref{sec:method}.

\newpage

\section{Additional Methodological Preliminaries}
This appendix section provides additional details on the learning and control concepts employed in this work. Specifically, we present the full optimization formulation of the MT-MPC in Section~\ref{subsec:apx_mtmpc}, describe the employed differentiable MPC scheme in Section~\ref{subsec:apx_diffmpc}, and summarize the basics of RNNs in Section~\ref{subsec:apx_rnn}.

\subsection{Multi-Trajectory Model Predictive Control}
\label{subsec:apx_mtmpc}
Under the assumption of sufficient computational resources and perfect environmental knowledge, the privileged teacher employs the structural idea of multi-trajectory model predictive control (MT-MPC) \citep{saccani2022multitrajectory}. The original formulation was motivated by safe navigation of unmanned aerial vehicles in unknown static environments perceived by an onboard extroceptive sensor. It guarantees that, provided initial feasibility, a feasible control sequence exists at every subsequent time step~$k$ that allows the agent to safely remain within the observable field of view, thereby ensuring safe operation under newly encountered conditions. This property is achieved by jointly predicting two pairs of state and input trajectories within the optimization: a safe trajectory pair and a performance trajectory pair, denoted with $[X_{k,s,N}^*, U_{k,s,N}^*]$ and $[X_{k,p,N}^*, U_{k,p,N}^*]$ respectively. While the safe trajectory remains within the field of view to guarantee safety, the performance trajectory aims to optimize the control objective. These trajectories are coupled through their first prediction step, via an equality constraint $x_{p,1} = x_{s,1}$, ensuring that the applied input $u_0^* = u_{s,0}^{*} = u_{p,0}^{*}$ balances safety and performance. A full MT-MPC formulation is given in the following:
\begin{equation}
\label{eq:mpc_mt}
\begin{aligned}
 &[X_{k,p,N_L}^*, U_{k,p,N_L}^*, X_{k,s,N_L}^*, U_{k,s,N_L}^*] := \argmin_{\substack{
\{x_{p,i}\}_{0}^{N_L+1}, \{u_{p,i}\}_{0}^{N_L} \\
\{x_{s,i}\}_{0}^{N_L+1}, \{u_{s,i}\}_{0}^{N_L}
}} \sum_{i=0}^{N_L} \ell(x_{p,i}, u_{p,i},c_i,r_i)  \\
    \text{s.t.} &\quad x_{p,0} = x(k), \quad x_{p,i+1} = f(x_{p,i},u_{p,i},\zeta_{d,i}), \quad x_{p,i} \in \mathcal{X}(\zeta_{e,i}), \quad u_{p,i} \in \mathcal{U}, \\
    &\quad x_{s,0} = x(k), \quad x_{s,i+1} = f(x_{s,i},u_{s,i},\zeta_{d,i}), \quad x_{s,i} \in \mathcal{X}(\zeta_{e,i}), \quad u_{s,i} \in \mathcal{U}, \\
    & \quad x_{s,N_L+1} \in \mathcal{F_{T}},\quad  x_{p,1} = x_{s,1}, \quad i=0,\hdots,N.
\end{aligned}
\end{equation}
Thereby $\mathcal{F_{T}}$ indicates a valid terminal set that ensures the agent stays within the field of view.

\subsection{Differentiable Model Predictive Control}
\label{subsec:apx_diffmpc}
The network inferring the objective parameters of our student MPC is trained by minimizing the loss in Eq.~\ref{eq:problem} using a gradient-based algorithm. Accordingly, we need to propagate gradients from the imitation loss $\mathcal{L}([X_{k,N_S}^{\theta,*}, U_{k,N_S}^{\theta,*}], [X_{k,p,N_L}^*, U_{k,p,N_L}^*])$, to the weights $\theta$ of our mapping function $\mathcal{R}(\theta,x(k),\underline{\boldsymbol{\hat{\zeta}}}_{d,\tilde{N}^d_k}(k),\underline{\boldsymbol{\hat{\zeta}}}_{e,\tilde{N}^e_k}(k))$, outputting the context-dependent cost parameters $\textbf{q}_{N_S}^{\theta}$. This propagation is described by the following equation:

\begin{equation}
\label{eq:chain_rule}
\frac{\partial \mathcal{L}}{\partial \theta} = \frac{\partial \mathcal{L}}{\partial [X_{k,N_S}^{\theta,*}, U_{k,N_S}^{\theta,*}]} \cdot \frac{\partial [X_{k,N_S}^{\theta,*}, U_{k,N_S}^{\theta,*}]}{\partial \textbf{q}^{\theta}_{N_S}} \cdot \frac{\partial  \textbf{q}^{\theta}_{N_S}}{\partial \theta}.
\end{equation}
However, propagating gradients through an optimization procedure, which in our case corresponds to the MPC problem of our student $\mathit{MPC}_{N_S}^{\theta}$, is not trivial~\cite{donti2017task, wilder2019melding, lahoud2024learning}.
Therefore, in accordance to~\cite{rickenbach2025zipmpc}, we leverage an existing differentiable MPC technique and employ the approach of \cite{amos2018differentiable}, as it is compatible with general MPC cost formulations. This method extends KKT-based differentiation \cite{amos2017optnet} to non-convex settings by iteratively solving convex quadratic approximations using the iterative Linear Quadratic Regulator (iLQR).

\subsection{Recurrent Neural Networks}
\label{subsec:apx_rnn}
To enable our parametric mapping function $\mathcal{R}(\theta,x(k),{\boldsymbol{\hat{\zeta}}}_{d,\tilde{N}^d_k}(k),{\boldsymbol{\hat{\zeta}}}_{e,\tilde{N}^e_k}(k))$ to retain memory, we employ a recurrent neural network (RNN)~\citep{rumelhart1986learning}. This class of neural network architectures processes sequential data by leveraging an internal hidden state that evolves over time, allowing the network to retrieve information from previous inputs, thereby giving it memory capabilities. Formally, given an input sequence $\{o_k\}_{k=0}^{T}$, in our case chosen as the padded perceived context of interest $\{o_k\}_{k=0}^{N_L}$, an RNN updates its hidden state $h_k$ recursively according to $h_k = \rho_{\theta}(o_k, h_{k-1}),$ where $o_k$ denotes the input at time step $k$, $h_{k-1}$ the hidden state from the previous time step, and $\rho_{\theta}$ a parameterized nonlinear mapping. The output of the network is then obtained as $y_k = g_{\theta}(h_k),$ where $g_{\theta}$ maps the hidden state to the desired output. Note that in the employed set-up $y_k$ corresponds to our objective parameter adaptation vector $\mathbf{q}_{N_S}^{\theta,k}$.\\
The hidden state thereby serves as an internal memory, allowing the network to retain relevant information from past observations.  In practice, variants such as long short-term memory (LSTM) \citep{graves2012long} and gated recurrent units (GRUs) \citep{cho2014learning} are commonly used to improve training stability and capture long-range dependencies. This property makes RNNs particularly suitable for partially observable settings, where relevant environmental information may no longer be directly available at the current time step.

\section{Detailed Imitation Learning Algorithm}
In this appendix section, we present the complete imitation learning algorithm in detail.
\paragraph{Detailed Algorithm.}

\begin{algorithm}[H]
    \caption{VIGOR-MPC}
    \textbf{Inputs:} \\
    Initial weights $\theta$ of the multi-level RNN $\mathcal{R}_\theta$.\\
    Horizon lengths $N_S$, $N_L$, and loss horizon $N_D \leq N_S$.\\
    Sequence length $T$.\\
    Cost parameters $C_{N_L}$ of the MT-MPC and nominal cost parameters $\mathbf{q}_{N_S}$ of $\mathit{MPC}_{N_S}^{\theta}$.\\
    $N_{E}$ simulation environments $\mathcal{O}_e$, $e=1,\hdots,N_{E}$.\\
    State-dependent field-of-view operator $\mathcal{F}_{W,x(k)}$.\\
    Number of training iterations per environment $N_I$.\\
    \vspace{-0.8em}
    \begin{algorithmic}[1]
    \label{alg:ml_mpc}
        \FOR{$e = 1$ to $N_{E}$}
            \FOR{$i = 1$ to $N_{I}$}
                \REPEAT
                    \STATE $x(k-T+1), \hdots, x(k) \gets \text{SAMPLESEQ}(\mathcal{O}_e, T)$
                \UNTIL{$\mathit{MPC}_{MT,N_L}$ is feasible at $x(t)$ for all $t = k-T+1,\hdots,k$}
                \STATE $H_{k-T} \gets \mathbf{0}$
                \FOR{$t = k-T+1$ to $k$}
                    \STATE $\hat{\mathbf{r}}_{\tilde{N}^r_t}(t), \hat{\boldsymbol{\zeta}}_{d,\tilde{N}^d_t}(t), \hat{\boldsymbol{\zeta}}_{e,\tilde{N}^e_t}(t) \gets \mathcal{F}_{W,x(t)}(\text{CONTEXT}_{\text{pc}}(\mathcal{O}_e,x(t)))$
                    \STATE $[\Delta \mathbf{q}_{N_S}^{\theta,t}, H_t] \gets \mathcal{R}_\theta(x(t), \hat{\boldsymbol{\zeta}}_{d,\tilde{N}^d_t}(t), \hat{\boldsymbol{\zeta}}_{e,\tilde{N}^e_t}(t), H_{t-1})$
                \ENDFOR
                \STATE $\mathbf{q}_{N_S}^{\theta,k} \gets \mathbf{q}_{N_S} + \Delta \mathbf{q}_{N_S}^{\theta,k}$
                \STATE $\mathbf{r}_{N_L}(k), \boldsymbol{\zeta}_{N_L}(k) \gets \text{CONTEXT}_{\text{gt}}(\mathcal{O}_e,x(k))$
                \STATE $[X_{k,p,N_L}^*, U_{k,p,N_L}^*] \gets \mathit{MPC}_{MT,N_L}(C_{N_L}, \mathbf{r}_{N_L}(k), \boldsymbol{\zeta}_{N_L}(k), x(k))$
                \STATE $[X_{k,N_S}^{\theta,*}, U_{k,N_S}^{\theta,*}] \gets \mathit{MPC}_{N_S}(\mathbf{q}_{N_S}^{\theta,k}, \hat{\mathbf{r}}_{\tilde{N}^r_k}(k), \hat{\boldsymbol{\zeta}}_{d,\tilde{N}^d_k}(k), \hat{\boldsymbol{\zeta}}_{e,\tilde{N}^e_k}(k), x(k))$
                \STATE $\mathcal{L} \gets \mathrm{MSE}_{N_D}([X_{k,N_S}^{\theta,*}, U_{k,N_S}^{\theta,*}], [X_{k,p,N_L}^*, U_{k,p,N_L}^*])$
                \STATE $\theta \gets \text{Adam}(\theta, \nabla_\theta \mathcal{L})$ \COMMENT{backprop through MPC and all $T$ frames}
            \ENDFOR
            \STATE Update simulation environment $\mathcal{O}_e$
        \ENDFOR
    \end{algorithmic}
\end{algorithm}

The step-by-step training of VIGOR-MPC using a gradient-based learning approach is presented in Algorithm \ref{alg:ml_mpc}. As inputs to the algorithm, we provide the structure of our memory-capable RNN, $\mathcal{R}$, its initial weights $\theta$, the horizon lengths of our perfectly knowledgeable teacher and our student, $N_L$ and $N_S$ respectively, a set of well-tuned cost parameters of our teacher MT-MPC, $C_{N_L}$, and a set of manually set initial cost parameters $\mathbf{q}_{N_S}$ of our student MPC. Furthermore, we provide $N_E$ simulation environments $\mathcal{O}_e$, with $e=1, \hdots,N_E$, that in the considered application example of an autonomous car would consist of randomly chosen obstacle configurations, as well as the camera-specific and state dependent field of view operator $\mathcal{F}_{W,x(k)}$. Additionally, we denote with $\text{CONTEXT}_{\text{gt}}$ and $\text{CONTEXT}_{\text{pc}}$ the ground truth and perception-derived context, respectively, and with $\text{SAMPLESEQ}$ we denote the sequence of states obtained by simulating the long horizon teacher control strategy for $T$ time steps from a randomly initialized initial state. To provide a wider range of context variation during training, we iterate over $N_{E}$ environmental setups (lines 1 and 13). For each environment, we iterate over several training steps $N_{I}$, in which we sample a batch of initial states $x(k)$ and compute the loss using empirical risk minimization over the batch.  
The manually set parameters of our student MPC $\mathbf{q}_{N_S}$ are used as a warm start for the learning algorithm, and the employed RNN outputs the correction values $\Delta \mathbf{q}_{N_S}^{\theta}$, which we add to $\mathbf{q}_{N_S}$ inferring the learned cost parameters $\mathbf{q}_{N_S}^{\theta}$ (lines 9 and 11). Finally, we adopt the mean squared error (MSE) as an imitation loss between the predicted states of the performance trajectory of the teacher MT-MPC and the trajectory of the student MPCs. In practice, we consider only the first $N_D \leq N_S$ steps in the imitation loss, i.e., $MSE_{N_D}$, and use Adam as the gradient-based optimizer to update the RNN weights (lines 13 to 16).

\section{Implementation Details}
\label{apx:implementation_details}
In the following, we provide all the necessary implementation details of running our example application case, the autonomous driving, in simulation. We provide the model and control details in Section~\ref{apx:examplesimsetup}, followed by the perception pipeline description in Section~\ref{apx:perception} and network and training details in Section~\ref{apx:learningframeworksetup}.

\subsection{Example Simulation Setup}
\label{apx:examplesimsetup}
In this section, we describe the employed coordinate system, the Frenet frame, the corresponding vehicle dynamics model, and the resulting MPC formulation used for racing. In addition, the selected model and control parameters are summarized.

\begin{figure}[htb]
\vspace{-0.5em}
    \centering
    \includegraphics[width=0.3\textwidth]{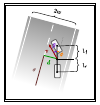}
    \caption{Illustration of the car's position in the Frenet frame, including its progress coordinate $\sigma$, its lateral deviation $d$, and its respective orientation $\phi$. Its velocity $v$ and the steering angle $\delta$ are also shown, as well as the car's rear and front length, $l_r$ and $l_f$.}
    \label{fig:frenetkin}
    \vspace{-1.0em}
\end{figure}

\paragraph{Frenet Frame}
The Frenet coordinate frame \cite{milliken1995race} is widely used in racing applications. It describes a reference-based coordinate system and hence defines the vehicle’s position relative to a given trajectory through its progress on a track or path, the lateral deviation from the corresponding center line, and the agent's orientation relative to the path, denoted by $\sigma$, $d$, and $\phi$, respectively. The corresponding position vector is defined as $x = [\sigma,d,\phi]$ and illustrated in Figure~\ref{fig:frenetkin}.

\paragraph{Car Dynamics}

To model the car's dynamics, we adopt the \emph{kinematic bicycle} model~\cite{rajamani2011vehicle}, detailed below. Thereby, $l_r$ denotes the distance from the rear axle to the center of mass, and $l_f$ the distance from the center of mass to the front axle. Additionally, we indicate the curvature with $\kappa(\sigma)$, which is progress-dependent. For an illustration of the introduced parameters, refer to Figure \ref{fig:frenetkin}.

The kinematic bicycle model assumes negligible tire slip and treats the wheels as rolling without lateral force generation. This makes it less accurate than dynamic models such as the \emph{Pacejka} tire model~\citep{pacejka2005tire} in high-speed or aggressive-maneuvering regimes, where slip angles become significant. However, for the low-to-moderate velocity range and acceleration profiles relevant to our F1TENTH platform, the kinematic model has been shown to track real vehicle behavior closely~\cite{kong2015kinematic}, while being substantially cheaper to evaluate inside the MPC optimization. This trade-off motivates its use both in our teacher and student controllers. The model takes the motor torque~$T_M$ and the front wheel steering angle~$\delta$ as control inputs. The side slip angle is defined as
\begin{align*}
    \beta(k) = \tan^{-1}(\frac{l_r}{l_f+l_r}\tan(\delta(k))).
\end{align*}
The discrete-time dynamics in Frenet frame, discretized using the Euler method with sampling time $T$, follow as
\begin{align*}
    &\sigma(k+1) =  \sigma(k) + T(v(k)\frac{\cos{(\phi(k) + \beta(k))}}{1 - \kappa(\sigma(k))d(k)}) \\
    &d(k+1) = d(k) +  T v(k)\sin({\phi(k) + \beta(k)}) \\
    &\phi(k+1) = \phi(k) +  T(\frac{v(k)}{l_f}\sin{(\beta(k))} - \kappa(\sigma(k))v(k) \frac{\cos{(\phi(k)+\beta(k))}}{1 - \kappa(\sigma(k))d(k)})\\
    & v(k+1) = v(k) + T \frac{1}{\tau}(a T_M(k) - v(k)).
\end{align*}
Note that the curvature~$\kappa(\sigma)$ appears explicitly in three of the four dynamic equations, reinforcing its role as the dominant environmental context shaping the vehicle's evolution along the track. Parameter values used in our experiments are listed in Table~\ref{tab:dyn_params}.

\begin{table}[h]
    \centering
    \caption{Powertrain and slip angle parameter values.}
    \label{tab:powertrain_params}
    \begin{tabular}{lccccc}
        \toprule
        \textbf{Parameter} & $l_r$ & $l_f$ & $a$ & $\tau$ \\
        \midrule
        \textbf{Value}     & 0.13429 & 0.12571 & 19.995 & 0.422 \\
        \bottomrule
    \end{tabular}
    \label{tab:dyn_params}
\end{table}

\begin{table}[h]
    \centering
    \caption{State and input constraint parameter values.}
    \label{tab:state_input_constraints}
    \begin{tabular}{lc}
        \toprule
        \textbf{Parameter} & \textbf{Value} \\
        \midrule
        $\sigma_{\min}$        & 0          \\
        $\sigma_{\max}$        & -         \\
        $\phi_{\min}$          & $-2$       \\
        $\phi_{\max}$          & 2          \\
        $v_{\min}$             & 0          \\
        $v_{\max}$             & 1.1--2.2   \\
        $T_{M,\min}$           & 0          \\
        $T_{M,\max}$           & 1.0        \\
        $\dot{T}_{M,\min}$     & $-15.0$    \\
        $\dot{T}_{M,\max}$     & 15.0       \\
        $\delta_{\min}$        & $-0.34$    \\
        $\delta_{\max}$        & 0.34       \\
        $\dot{\delta}_{\min}$  & $-2.0$     \\
        $\dot{\delta}_{\max}$  & 2.0        \\
        $\omega$               & 0.457      \\
        \bottomrule
    \end{tabular}
    \label{tbl:model_params}
\end{table}

\subsection{Perception Pipeline Details}
\label{apx:perception}

The perception module processes either onboard RGB-D data or simulated onboard RGB-D images rendered in Blender to extract the centerline reference $\hat{\mathbf{r}}_{\tilde{N}^r_k}(k)$ and track boundary context $\hat{\boldsymbol{\zeta}}_{e,\tilde{N}^e_k}(k)$ used by the controller. The pipeline consists of three main stages: depth-based mask filtering, edge extraction via Edge Drawing~\citep{topal2012edge}, and centerline estimation from the detected lane markings. 

\subsubsection{Depth-Based Mask Filtering}
\label{app:perception:mask}

To restrict subsequent processing to the drivable floor region, we generate a binary mask that suppresses obstacles, reflections, and background clutter. Given a camera-to-floor transformation $(R_{CC'},\, {}^{C'}r_{CC'})$, we synthesize an \emph{artificial depth map} $D_{\mathrm{art}}$ representing the expected depth of the floor at every pixel up to a fixed horizon distance. The smart mask is then obtained by comparing the measured depth map $D_{\mathrm{meas}}$ against $D_{\mathrm{art}}$:
\begin{itemize}
    \item pixels with $D_{\mathrm{meas}} \ll D_{\mathrm{art}}$ are classified as \emph{obstacles},
    \item pixels with $D_{\mathrm{meas}} \gg D_{\mathrm{art}}$ are classified as \emph{reflections},
    \item pixels with invalid (\texttt{NaN}) depth are discarded.
\end{itemize}
The comparison uses a multiplicative threshold that loosens with distance to account for the growing stereo depth uncertainty, with a tighter lower bound (obstacles) than upper bound (reflections), as obstacles are both easier to identify and more critical to exclude. The resulting mask restricts all downstream processing to pixels that lie on the expected floor surface.

\subsubsection{Lane Marking Detection via Edge Drawing}
\label{app:perception:edge}

Within the masked floor region, lane markings are extracted using the Edge Drawing algorithm~\cite{topal2012edge}. After converting the color image to grayscale and applying Gaussian smoothing, horizontal and vertical gradients are computed with Sobel filters and thresholded to retain only strong responses. Edge Drawing then identifies \emph{anchor points} at local maxima of the gradient magnitude and traces edges by following the highest-gradient neighbors. The smart mask from Section~\ref{app:perception:mask} is applied to filter out anchors lying on obstacles or reflections, both improving accuracy and reducing the tracing cost. As a post-processing step, edges extending beyond the horizon distance and edges shorter than a minimum length are discarded. Unlike conventional edge detectors that return a binary edge image, Edge Drawing returns each edge as an ordered, contiguous sequence of pixels with a well-defined start and end, which is essential for the subsequent centerline extraction.

\subsubsection{Centerline Estimation from Lane Markings}
\label{app:perception:centerline}

The detected edge pixels are projected onto the calibrated floor plane to obtain a metric 2D representation of the lane markings. Each projected edge is smoothed by a moving average over its floor coordinates to attenuate pixelation noise, sparsified by enforcing a minimum spacing of $2$--$4$\,cm between consecutive points, and oriented such that the point closest to the camera origin is its starting point.

The processed edges are then grouped into lane markings using two criteria: spatial \emph{proximity} and \emph{orientation consistency}. Proximity is evaluated only from the first and last points of each edge to reduce computational cost, using a threshold related to the lane-marking thickness. Orientation consistency is evaluated via the dot product of the edge direction vectors, each defined from the corresponding edge endpoints, and used to reorient edges within a group when needed, with the edge closest to the camera origin serving as the orientation reference. Edges that fail either the proximity or orientation criteria are discarded. To avoid redundant comparisons, a visited-edge map is maintained such that each edge is processed only once. For each grouped lane marking, a \emph{sidedness} label in $\{-1, +1\}$ is assigned by projecting the inter-edge vector onto the lane-marking normal, indicating whether the marking lies to the left or right of its neighbor.

Finally, the centerline is computed by pairing lane markings of opposite sidedness within the same group. For paired markings, midpoints are obtained by an interleaved, recursively refined moving average over corresponding edge points, leveraging the ordered structure of the edge samples to establish local point correspondences. The recursion deepens whenever the connecting vector between matched points deviates beyond an angular threshold, ensuring robustness to local mismatches in sampling, and continues until all points are processed or a maximum recursion depth is reached. For unpaired segments, where only one of the two lane markings is visible, the assumption that the ego vehicle remains within the lane uniquely determines the inward offset direction, and the centerline is reconstructed by offsetting the visible marking by the known half-lane width $\omega$ along its inward normal. For partially overlapping lane markings, offsets are applied only to unmatched residual sections after recursive pairing. The concatenated midpoints form the centerline estimate $\hat{\mathbf{r}}_{\tilde{N}^r_k}(k)$ passed to the controller, while the grouped lane markings provide the track-boundary context $\hat{\boldsymbol{\zeta}}_{e,\tilde{N}^e_k}(k)$.

\subsubsection{Obstacle Detection}
\label{app:obstacle}

Following the centerline extraction of Section~\ref{app:perception:centerline}, we detect the obstacles and convert them into a progress-dependent narrowing of
the admissible lateral corridor. The result is forwarded directly as environmental context $\hat{\boldsymbol{\zeta}}_{e,\tilde{N}^e_k}(k)$ and the boundary
constraint $-\omega_l(\sigma_i) \le d_i \le \omega_r(\sigma_i)$ of the trained short-horizon MPC. In our example setup, the obstacles are
marked in a distinct color, so detection reduces to a color-segmentation problem.
However, the geometric reasoning that follows is agnostic to how the obstacle
silhouette in image space is obtained.

\paragraph{Color segmentation and clustering.}
The RGB image is converted to the HSV color space and thresholded to a binary
mask isolating the obstacle color (in our case, yellow). Morphological opening and closing remove
speckle and fill small holes, and part  of the image above the horizon is masked out. The resulting mask is eroded
and decomposed into connected components, with components whose pixel area falls below a
minimum threshold discarded as noise. For each retained component~$j$ we
compute its convex hull, and collect
the set of its vertices.

\begin{figure}
    \centering
    \includegraphics[width=0.98\linewidth]{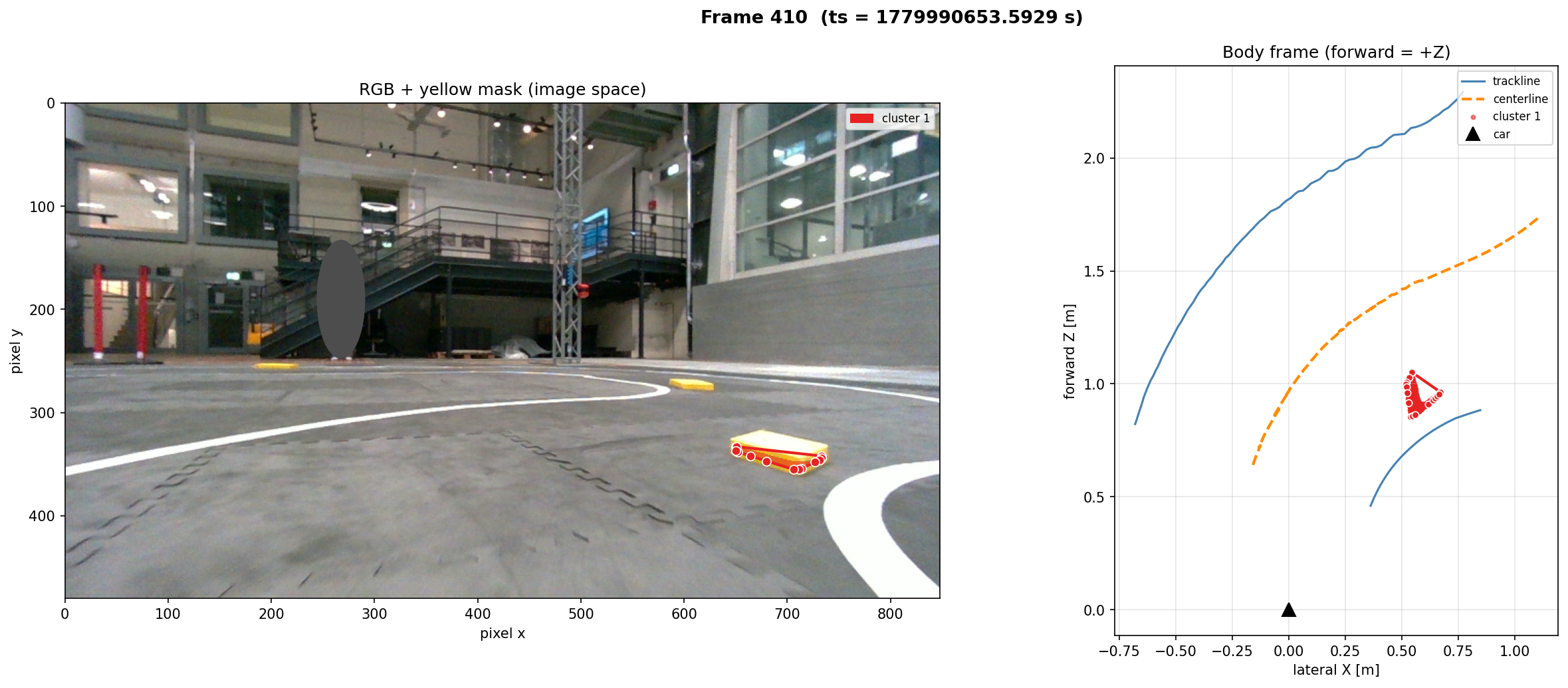}
    \caption{Example of an egocentric view from the car and the obstacle detection.}
    \label{fig:obst_detection}
\end{figure}

\paragraph{Unprojection to the body frame.}
Using the aligned depth image together with the camera calibration matrix~$K$ and the
extrinsic calibration $(R, T)$, the vertices of the convex hull of each cluster are
projected onto the ground plane, giving a set of body-frame points
$\{p_{j,m}\}_{m} \subset \mathbb{R}^2$ for cluster~$j$. Crucially, the centerline is
extracted in the {same} body frame and from the {same} image. The
curvilinear coordinates introduced below are therefore computed entirely in the
body frame, with the vehicle at the origin, as shown in Figure~\ref{fig:obst_detection}.

\paragraph{Curvilinear projection.}
Let $\{c_l\}_{l=0}^{L}$ denote the extracted centerline polyline in the body frame,
obtained after outlier rejection and one-dimensional Gaussian smoothing, and let
$\sigma_l$ be its cumulative arc length. The arc length is offset by the distance
from the vehicle to the first centerline point, $\sigma_0 = \lVert c_0 \rVert$, so
that the progress coordinate is measured consistently with the controller's
convention (the vehicle's own projection onto the centerline). For an obstacle
point $p$ we locate the foot of the normal onto the polyline,
\begin{equation}
  l^\star = \arg\min_{l} \operatorname{dist}\!\big(p,\, \overline{c_l\, c_{l+1}}\big),
  \qquad
  c^\star = \operatorname{proj}_{\overline{c_{l^\star} c_{l^\star+1}}}(p),
\end{equation}
and obtain its progress as $\sigma(p) = \sigma_{l^\star} + \lVert c^\star - c_{l^\star} \rVert$.

\paragraph{Track width adjustment.}
Each cluster~$j$ is summarized by the arc-length interval it occupies and by its
closest lateral approach to the centerline on each side. Considering only points
that lie within the nominal corridor, $|d| \le \omega$, we define
\begin{align}
  \sigma_j^{\min} &= \min_{m} \sigma(p_{j,m}), &
  \sigma_j^{\max} &= \max_{m} \sigma(p_{j,m}), \\[2pt]
  \underline{d}^{\,-}_j &= \!\!\min_{m:\, d(p_{j,m}) < 0}\!\! \big| d(p_{j,m}) \big|, &
  \underline{d}^{\,+}_j &= \!\!\min_{m:\, d(p_{j,m}) \ge 0}\!\! \big| d(p_{j,m}) \big| ,
\end{align}
i.e.\ the smallest absolute lateral distance attained by the cluster on the left
($d<0$) and right ($d \ge 0$) side, respectively. Starting from the nominal
half-width, $\omega_l(\sigma) = \omega_r(\sigma) = \omega$, every cluster tightens
the corresponding boundary over its arc-length span,
\begin{equation}
  \omega_l(\sigma) \leftarrow \min\!\big(\omega_l(\sigma),\, \underline{d}^{\,-}_j\big),
  \qquad
  \omega_r(\sigma) \leftarrow \min\!\big(\omega_r(\sigma),\, \underline{d}^{\,+}_j\big),
  \qquad \forall\, \sigma \in [\sigma_j^{\min}, \sigma_j^{\max}] .
\end{equation}
When several clusters overlap in arc length, the tightest bound prevails. The
outcome is a pair of progress-dependent corridor bounds $\omega_l(\sigma)$ and
$\omega_r(\sigma)$ that coincide with the nominal track half-width wherever no
obstacle is observed and shrink to the obstacle's nearest edge wherever one is
detected. 

\subsection{Learning Framework Setup and Implementation Details}
\label{apx:learningframeworksetup}

In the following, we detail the employed learning framework including some technical details on the available compute, the structure of the chosen neural network, and details on the training set-up. 

\paragraph{Technical Details.} Training was performed using 80 CPU cores, each allocated 4 GB of RAM (320 GB total memory). The generation of target trajectories via simulation was parallelized, and model training was carried out in mini-batches with a batch size of 80 over 20 epochs. {For execution in simulation an Intel(R) Core(TM) i7-13700K processor (13th generation), comprising 24 cores was employed.}

\paragraph{Input to the Neural Network.}
The RNN input consists of the available environmental context within the field of view, $\zeta_{k,N_L}$, together with a subset of state variables from $x(k)$. Note that to ensure generalizability and avoid overfitting to a specific training track, the track progress variable $\sigma$ was never provided to the RNN as state information, rather we use only $v$, $d$, and $\phi$ as RNN inputs, capturing the relevant local vehicle dynamics while remaining sufficiently track-independent. The dynamics context vector $\boldsymbol{\zeta}_{d,N_k}(k)$ is chosen as in~\cite{rickenbach2025zipmpc} and contains a sequence of track curvature values. Additionally, we augment the input with environmental context $\boldsymbol{\zeta}_{e,N_k}(k)$ that contains track boundary information on both sides, overlayed by detected obstacle information that tighten the track from the respective side. All context values are extracted from the current state onward and masked by the state dependent field-of-view operator $\mathcal{F}_{W,x(k)}$. Specifically, the sequence length is chosen to cover the maximum possible progress $\sigma{\Delta}$ over the MT-MPC prediction, which has been obtained by using the maximum velocity and progressing the context information forward for $N_L$ steps, before masking it with $\mathcal{F}_{W,x(k)}$. Missing or occluded context information is replaced with a noninformative padding value and is related to a negative mask value.

\paragraph{Neural Network and Training.}
The network input is divided into two components: (i) a global context vector containing three state-related features, and (ii) a variable-length sequential context representation encoding the perceived environmental information over the visible horizon.
To process the sequential input, we employ a single-layer gated recurrent unit (GRU) with hidden size $32$. Since the visible context length may vary between samples due to the limited field of view, the sequential inputs are padded and processed using packed sequences, with the corresponding valid lengths determined through a binary visibility mask. The GRU output sequence is subsequently aggregated through an attention-based pooling layer, yielding a fixed-dimensional latent representation that summarizes the available contextual information.
The pooled recurrent feature is concatenated with the global context vector and passed through a feedforward network consisting of six fully connected layers with $512$ hidden units each. All hidden layers use a LeakyReLU activation with slope $0.1$. Based on the resulting shared latent representation, the network predicts the MPC cost parameters through two separate output heads.
This architecture allows the student to combine short-term memory of previously observed environmental context with a structured prediction of horizon-dependent MPC objective parameters. Further details can be found in the code provided in the supplementary material.

\paragraph{Imitation Loss and Validation.}
As described in the main text, the imitation loss is defined as the sum of the mean squared error (MSE) between the student and teacher trajectories for both states and control inputs. Specifically, the loss is computed between the predicted trajectories of VIGOR-MPC and those of the privileged teacher $\mathit{MPC}_{MT,N_L}$. In line with the implementation details presented in \citep{rickenbach2025zipmpc}, we apply manually selected weighting factors to balance the contribution of the individual state and input dimensions and prevent individual variables from dominating the loss. The exact weights are provided in the supplementary code.
Furthermore, we also evaluate the loss only over the first $N_D$ prediction steps.
For model selection, lap performance on the training track is periodically evaluated from a fixed initial state throughout training. After completion, the model achieving the best lap performance on this validation rollout is selected.

\section{Hardware and Experimental Setup}
\label{apx:hardware_setup}

\subsection{Hardware Setup: Car and Camera}
For our real-world experiments we employ a modified F1Tenth platform~\citep{o2020f1tenth}
running the software stack of~\cite{carron2023chronos}. All computation is performed
onboard a LattePanda Sigma equipped with an Intel Core i5-1340P processor, so that both
perception and control run locally without any external compute. The vehicle carries an
Intel RealSense D405 RGB-D camera as its sole exteroceptive sensor, providing aligned
color and depth streams. In our experiments, we used $848 \times 480$ resolution and the rate of $90$\,FPS.

\subsection{Environment Setup: Track and Obstacles}
The track is assembled from interlocking foam mats and has a lane width of $1.10$\,m. Its
boundaries are marked with white adhesive tape on a black surface, providing the high
contrast required by the perception pipeline. Obstacles are introduced as physical objects
placed on the track, which the vehicle must perceive and avoid online. The corresponding
detection module is detailed in Appendix~\ref{apx:perception}.

\subsection{Experiment Setup}
The MPC is solved with acados~\citep{verschueren2022acados} at a control rate of $30$\,Hz.
From the RealSense color and depth streams, the perception pipeline detects the lane
markings and estimates the centerline, while obstacles are handled by the separate
detection module described in
Appendix~\ref{apx:perception}. Prior to each run, a calibration step uses an AprilTag
placed on the floor to recover the pose of the camera relative to the ground plane.
Ground-truth state estimates, which the controller assumes to be known, are obtained from
an external motion-capture system.

\section{Additional Results}
\label{apx:additional_results}
In this section, we present additional results, augmenting those in the main paper. Therefore, we provide additional simulation results in Section~\ref{subsec:add_sim_results}, and additional real-world results in  Section~\ref{subsec:add_hard_results}. This is then concluded with some visualizations of the training behaviour in Section~\ref{subsec:add_train_results}.

\begin {figure}[t]
\centering
\begin{tikzpicture}[scale=1.0]
   \node (pic1) at (-2.4, 0.0) {\includegraphics[width=.33\textwidth]{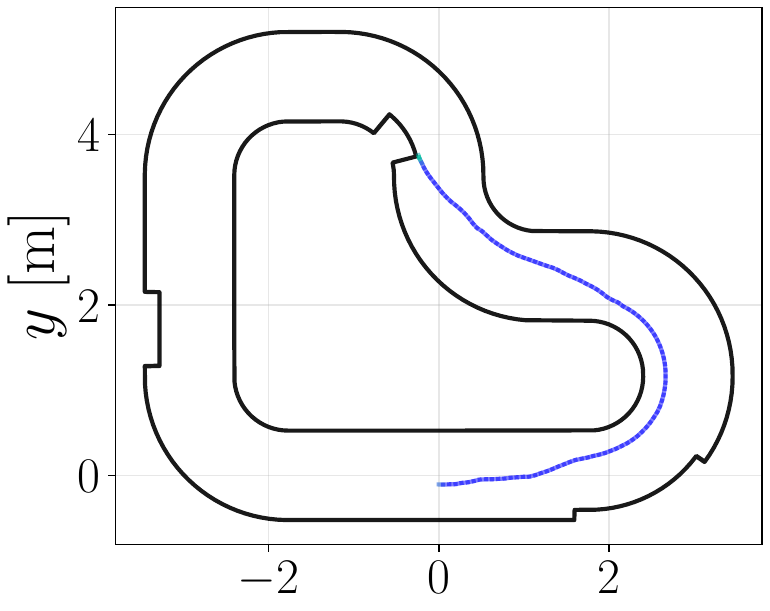}};
\node (pic2) at (2.25, 0.0) {\includegraphics[width=.305\textwidth]{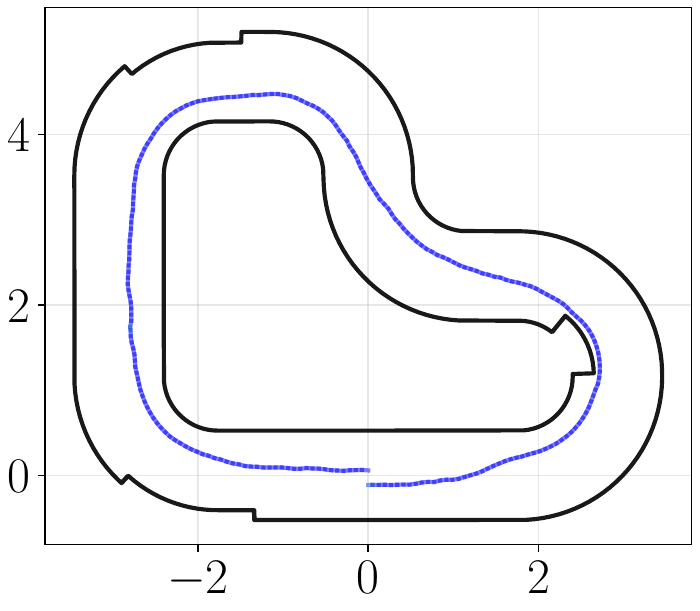}};
\node (pic3) at (6.8, 0.0) {\includegraphics[width=.305\textwidth]{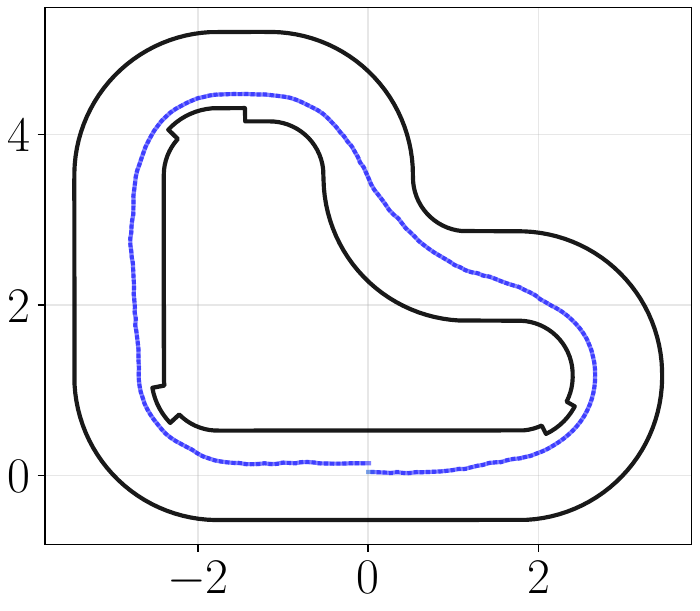}};
\node (pic4) at (-2.4, -3.8) {\includegraphics[width=.33\textwidth]{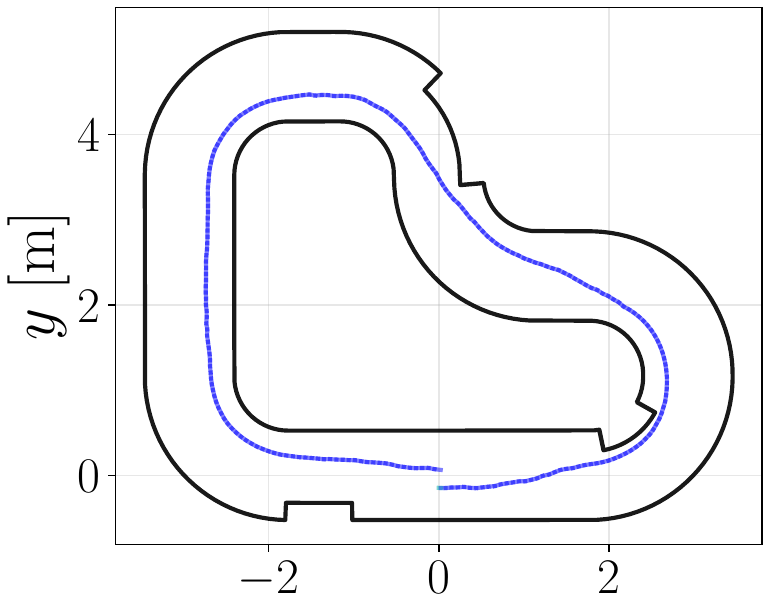}};
\node (pic5) at (2.25, -3.8) {\includegraphics[width=.305\textwidth]{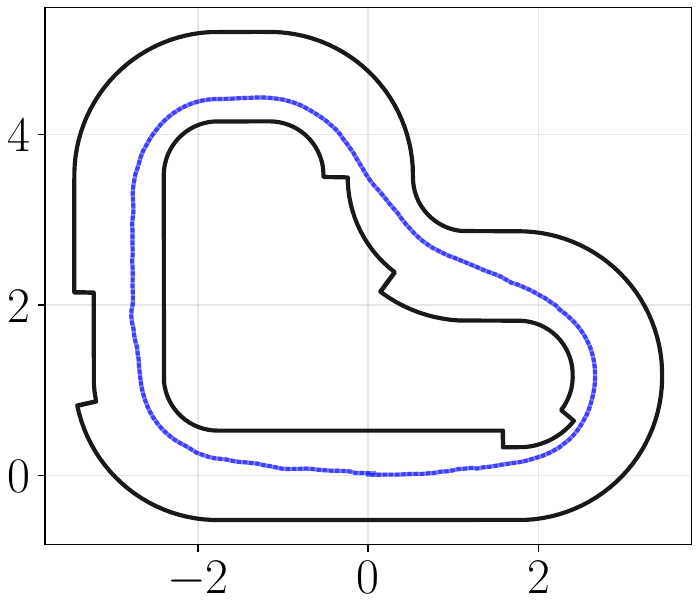}};
\node (pic6) at (6.8, -3.8) {\includegraphics[width=.305\textwidth]{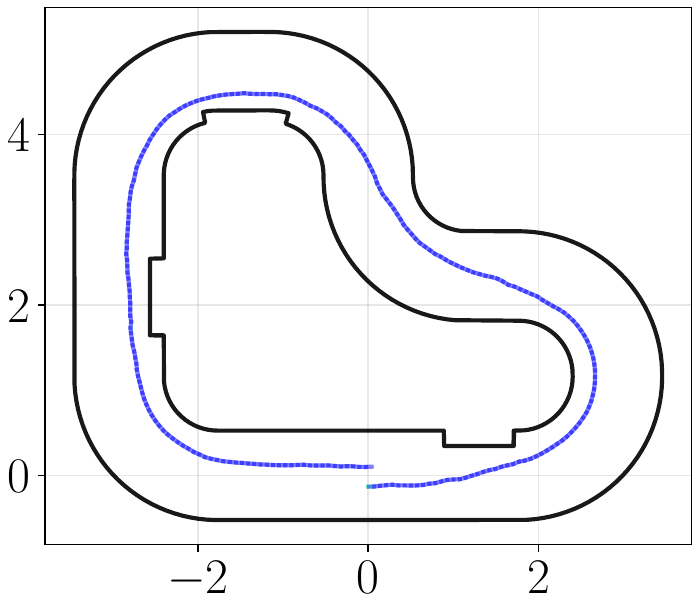}};
\node (pic7) at (-2.4, -7.8) {\includegraphics[width=.335\textwidth]{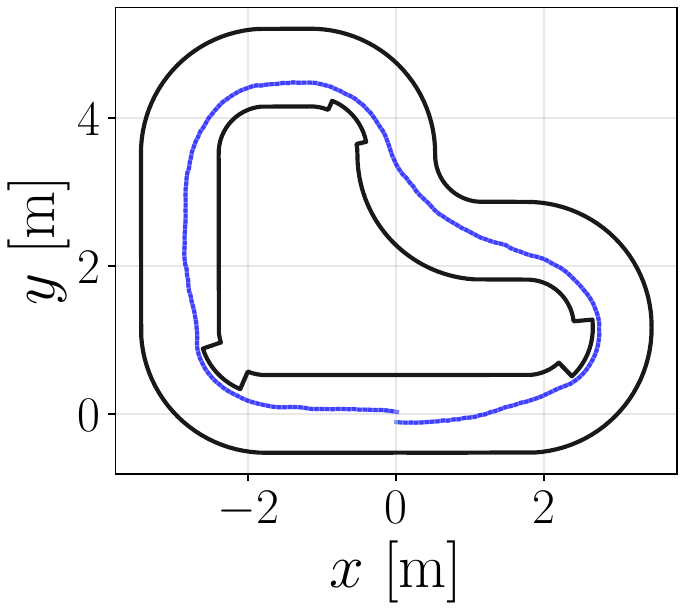}};
\node (pic8) at (2.25, -7.8) {\includegraphics[width=.305\textwidth]{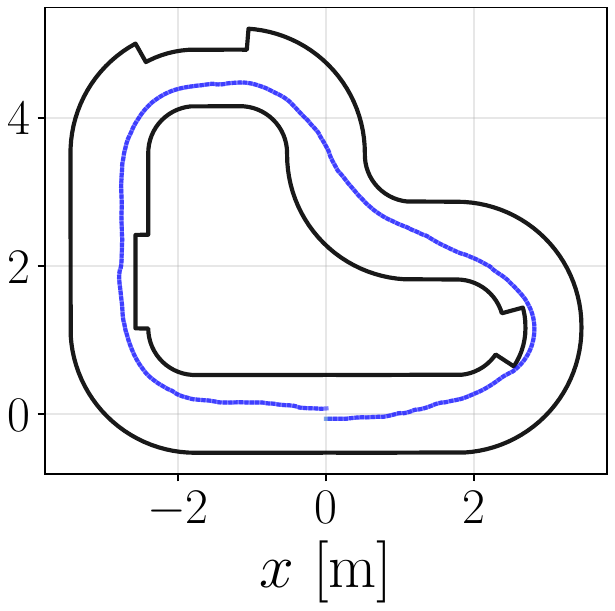}};
\node (pic9) at (6.8, -7.8) {\includegraphics[width=.305\textwidth]{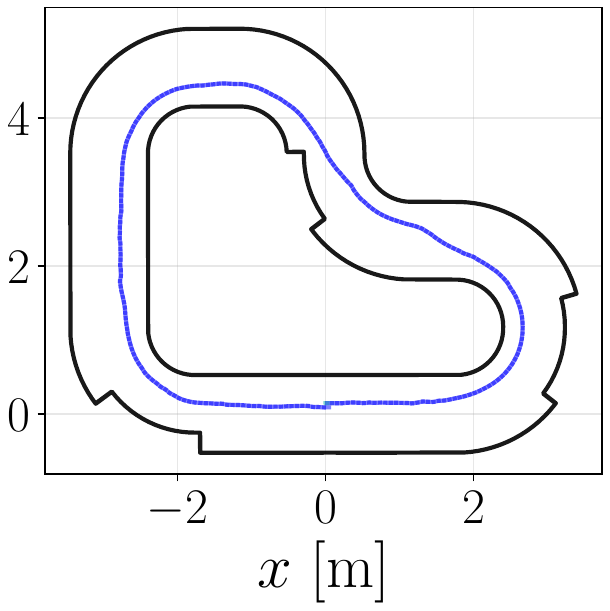}};

 \node (pic2) at (2.2, 3.3) {\includegraphics[width=.4\textwidth]{colorbar.pdf}};

\end{tikzpicture}
    \caption{{Closed-loop employment of VIGOR-MPC on different obstacle layouts (with randomly placed obstacle configurations).}} 
    \vspace{-1em}
\label{fig:addtnl_results}
\end{figure}

\subsection{Additional Simulation Results}
\label{subsec:add_sim_results}
In the following, we provide further {immitation error statistics,} visualization of the obstacle avoiding capabilities of VIGOR-MPC,{ a comparison with an end-to-end learned algorithm, and illustrate the effect of teacher horizon length on lap time and per-step solve time.}

{\textbf{Extended Heavy-Tail Statistics for Imitation Error:} To account for heavy-tailed distributions in Table~\ref{tbl:results_im_time_comp} of the paper, in Table~\ref{tab:additional_stats} we report the median, IQR, and p90. VIGOR-MPC reduces the median loss by 51.4\% relative to $\mathit{MPC_{N_S}}$, and substantially reduces the upper-tail error. Its p90 (tail performance) is comparable to the teacher baseline. The difference to $\mathit{MPC_{N_S}}$ is significant under a Wilcoxon signed-rank test ($p=7.6\cdot^{-12}$), which does not rely on a normality assumption.}

\begin{table}[t]
\centering
\caption{{Heavy-tail statistics of the imitation error: median, interquartile range (IQR), and 90th percentile (p90).}}
\label{tab:additional_stats}
\begin{tabular}{lccc}
\toprule
Method & Median & IQR & p90 \\
\midrule
VIGOR-MPC          & $0.00320$ & $[0.00055,\, 0.01341]$ & $0.02913$ \\
$MPC_{MT^{*},N_L}$ & $0.00152$ & $[0.00023,\, 0.00947]$ & $0.02871$ \\
$MPC_{N_S}$        & $0.00657$ & $[0.00107,\, 0.03552]$ & $0.05999$ \\
\bottomrule
\end{tabular}
\end{table}

\begin {figure}[H]
\centering
\begin{tikzpicture}[scale=1.0]
   \node (pic1) at (-2.4, 0.0) {\includegraphics[width=.355\textwidth]{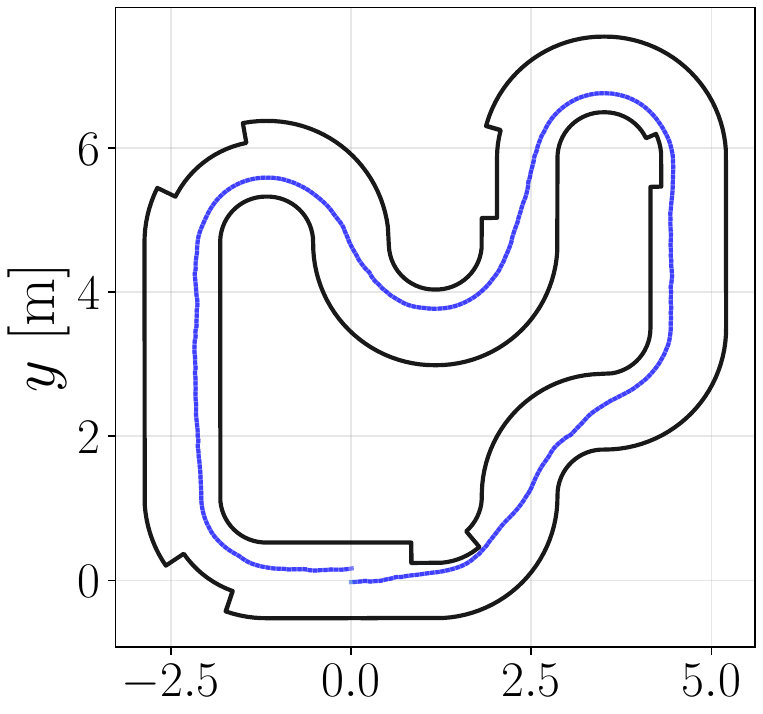}};
\node (pic2) at (2.3, 0.0) {\includegraphics[width=.32\textwidth]{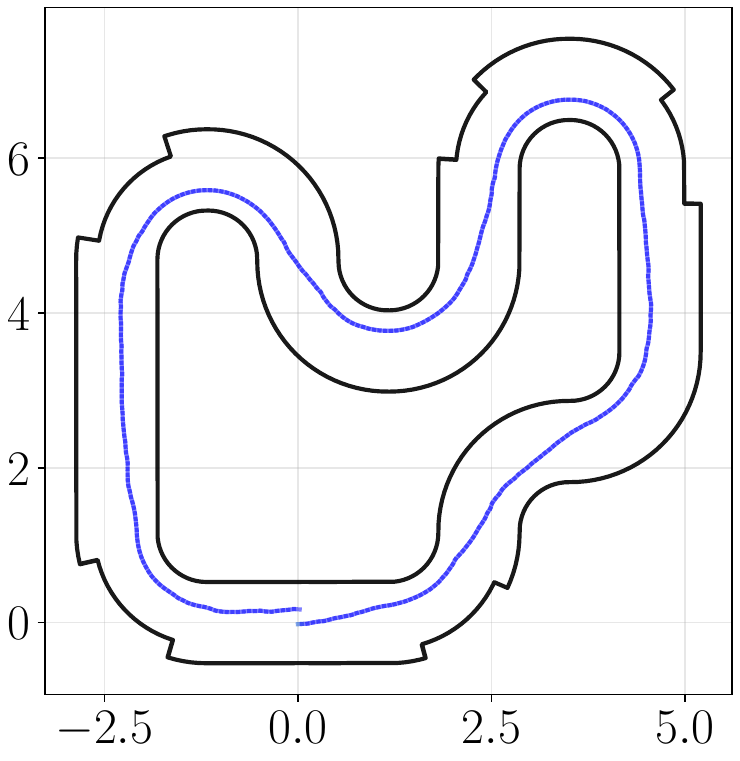}};
\node (pic3) at (6.8, 0.0) {\includegraphics[width=.32\textwidth]{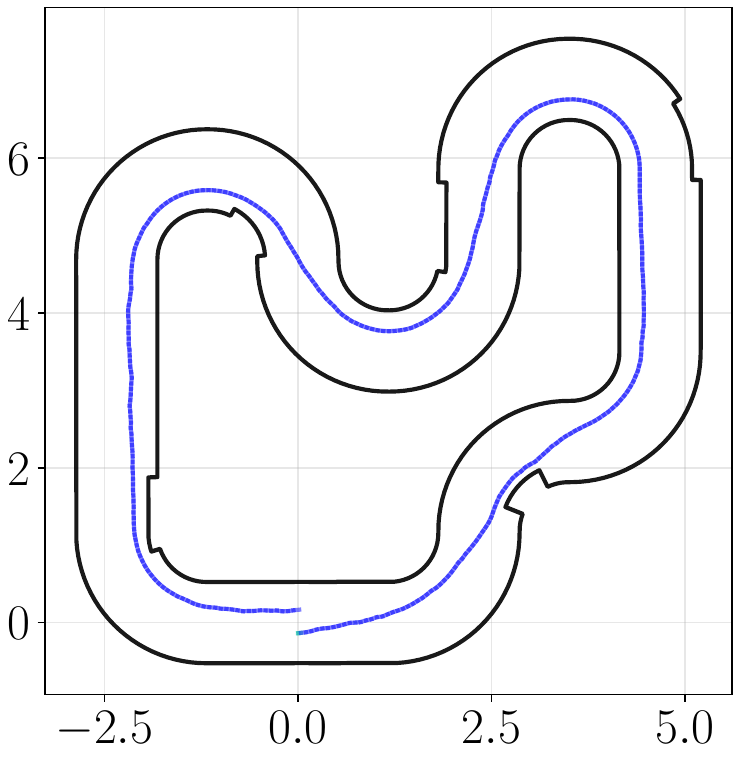}};
\node (pic4) at (-2.4, -4.7) {\includegraphics[width=.355\textwidth]{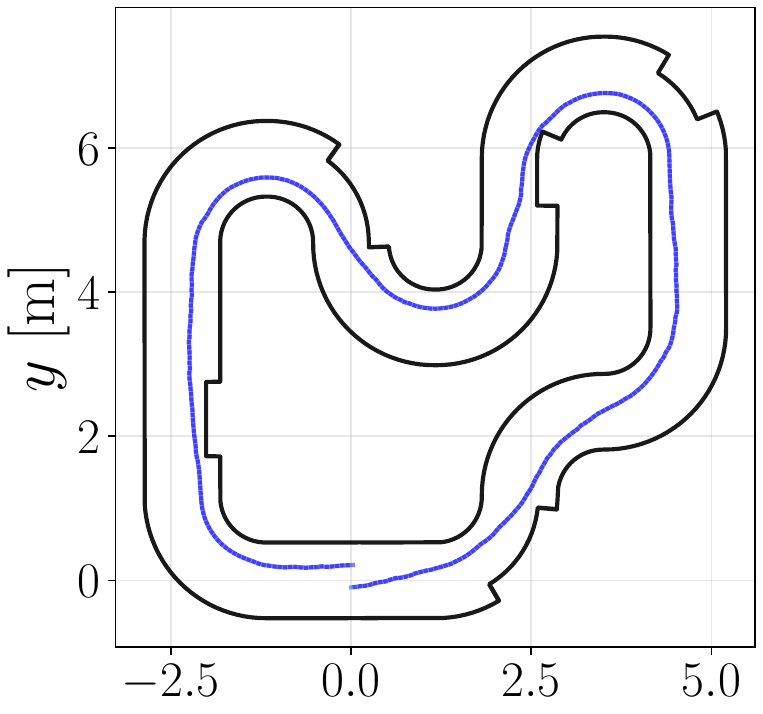}};
\node (pic5) at (2.3, -4.7) {\includegraphics[width=.32\textwidth]{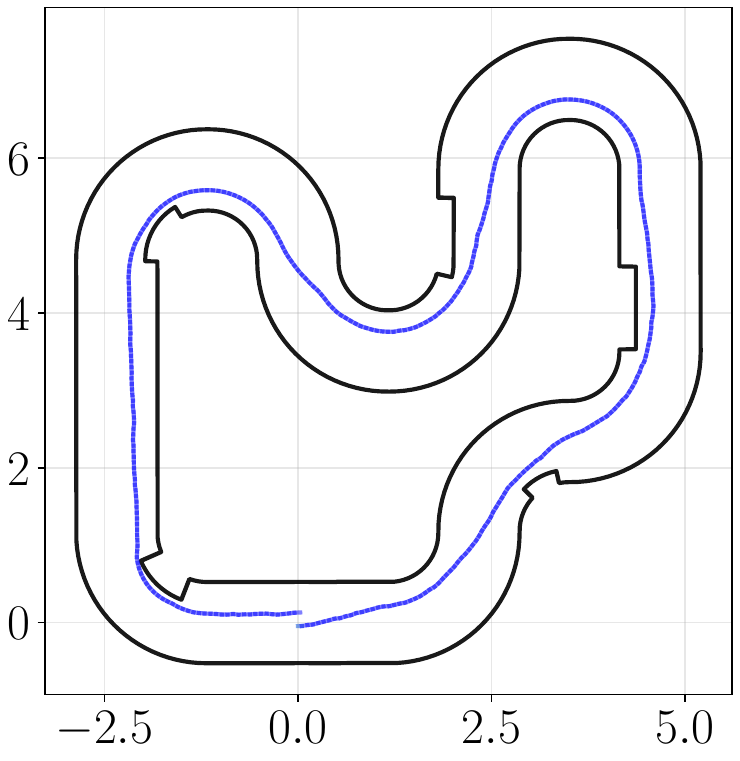}};
\node (pic6) at (6.8, -4.7) {\includegraphics[width=.32\textwidth]{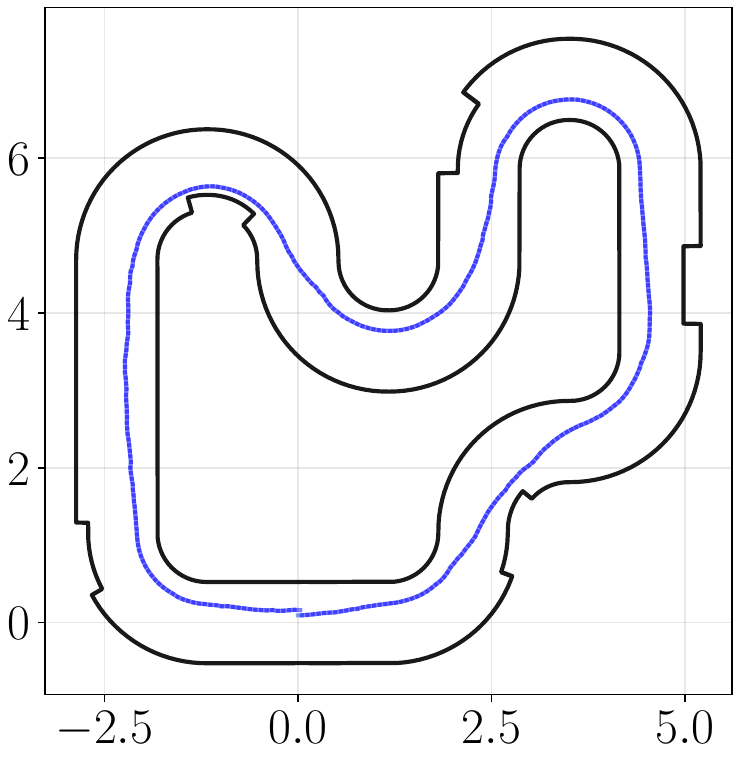}};
\node (pic7) at (-2.4, -9.6) {\includegraphics[width=.355\textwidth]{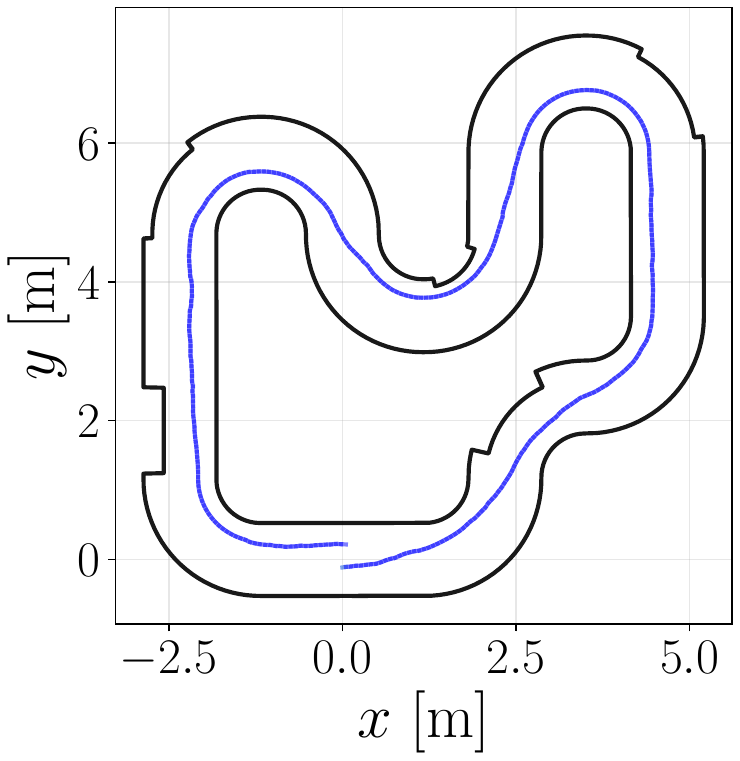}};
\node (pic8) at (2.3, -9.6) {\includegraphics[width=.32\textwidth]{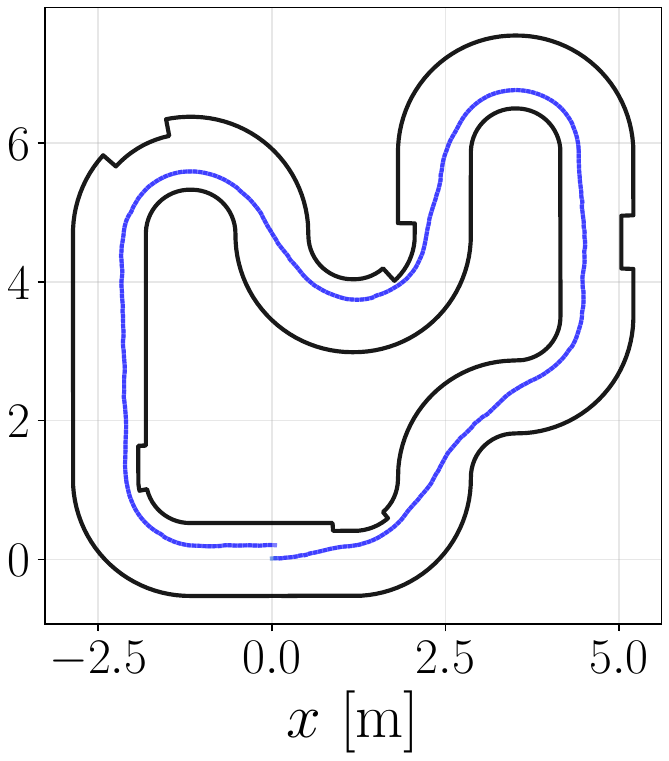}};
\node (pic9) at (6.8, -9.6) {\includegraphics[width=.32\textwidth]{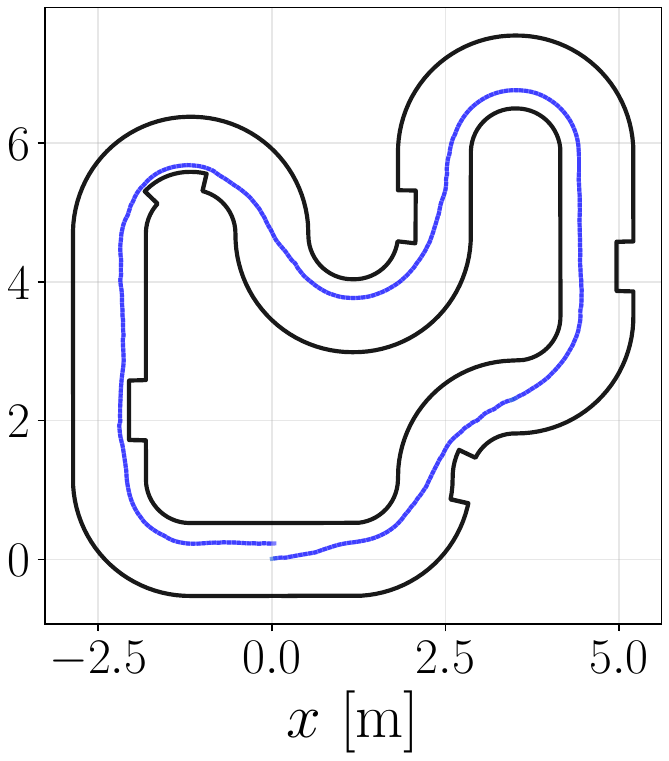}};
 \node (pic2) at (2.2, 3.4) {\includegraphics[width=.4\textwidth]{colorbar.pdf}};

\end{tikzpicture}
    \caption{{Closed-loop employment of VIGOR-MPC on different obstacle layouts (with randomly placed obstacle configurations) of  a track not seen during training.}} 
    \vspace{-1em}
\label{fig:addtnl_results1}
\end{figure} 

\begin {figure}[H]
\centering
\begin{tikzpicture}[scale=1.0]
   \node (pic1) at (-2.4, 0.0) {\includegraphics[width=.33\textwidth]{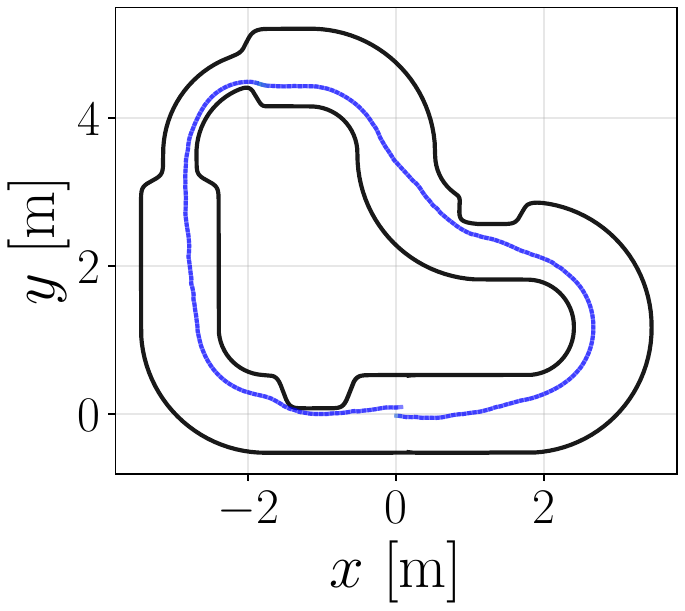}};
\node (pic2) at (2.25, 0.0) {\includegraphics[width=.305\textwidth]{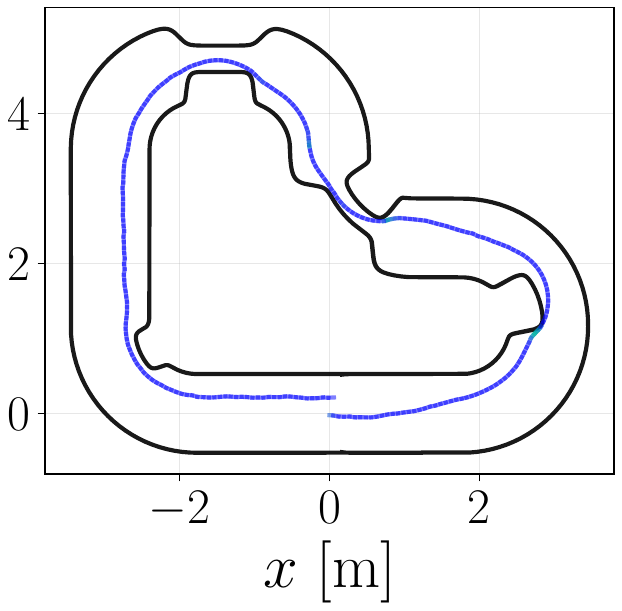}};
\node (pic3) at (6.8, 0.0) {\includegraphics[width=.305\textwidth]{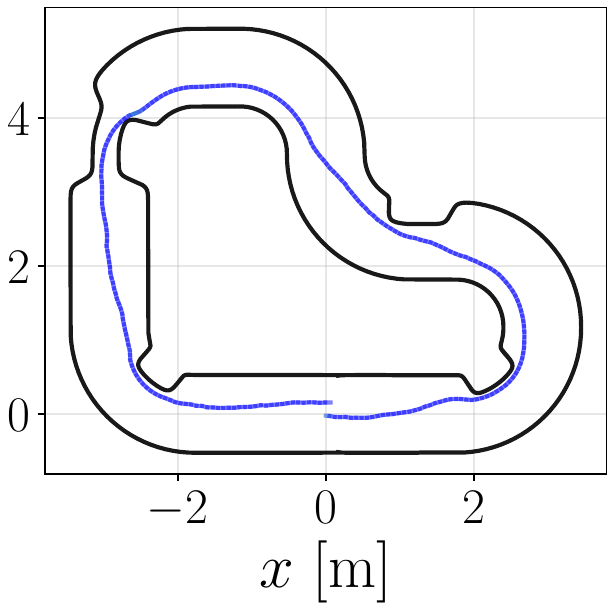}};

 \node (pic2) at (2.2, 3.3) {\includegraphics[width=.4\textwidth]{colorbar.pdf}};

\end{tikzpicture}
    \caption{{Closed-loop employment of VIGOR-MPC on different obstacle layouts (with intentionally placed obstacle configurations).}} 
    \vspace{-1em}
\label{fig:addtnl_results_hard}
\end{figure}

\textbf{Further Results on Obstacle Avoidance:} The following results augment Figure~\ref{fig:obstacle_avoidance} with additional visualizations of successfully completed laps under randomly generated obstacle configurations. We also include the single configuration, out of the 100 tested, in which VIGOR-MPC failed to complete the lap. Notably, even in this case it came to a safe stop rather than crashing into the obstacle. In Figure~\ref{fig:addtnl_results1} we show how VIGOR-MPC, trained only on the left track of Figure~\ref{fig:generalization}, avoids obstacles on configurations of a track not seen during training. Across 100 different obstacle configurations, it never crashes and always completes the lap successfully. Finally, in
Figure~\ref{fig:addtnl_results_hard}, we provide examples of obstacle
configurations intentionally placed to be difficult for the car to navigate. None
of these were seen during training, not even the general placement patterns, such
as obstacles on both sides of the track at the same arc length. Despite this,
VIGOR-MPC avoids the obstacles and completes the laps in closed loop.

\textbf{Comparison with an End-to-End Learned Policy:}
To emhasize the importance of the  MPC controller structure when it comes to ensuring constraint satisfaction, we compare VIGOR-MPC against an end-to-end policy
that imitates the same teacher but predicts control inputs directly. This baseline
receives inputs identical to those of VIGOR-MPC, namely the global state features
$v$, $d$, $\phi$ together with the field-of-view-limited horizon of track curvature
and left and right boundary distances, and it shares the same recurrent backbone
formed by a GRU encoder, attention pooling over the visible context, and a fully
connected head. The only structural difference lies in the output and what follows
it. Whereas VIGOR-MPC emits cost-parameter adjustments used in
short-horizon MPC, which is then solved as a differentiable layer to produce the
applied control, the baseline maps the shared representation directly to the two
control inputs, torque and steering.

To ensure that any performance gap cannot be attributed to insufficient capacity,
we deliberately grant the baseline a strictly larger backbone. We deepen the
recurrent encoder from one to three GRU layers and double its hidden state from
$32$ to $64$ units, while keeping the six fully connected layers of width $512$
unchanged. The policy is trained on the same data and for the same number of
epochs, using a weighted mean-squared error that regresses its output onto the
first applied control of the long-horizon MT-MPC teacher. 

Despite its larger capacity and identical supervision, the end-to-end policy
exhibits substantially worse constraint satisfaction. Evaluated on the same $100$ obstacle
configurations used to assess VIGOR-MPC, it crashed into an obstacle or the track
boundary in $89$ of the $100$ runs. Representative laps are shown in
Figure~\ref{fig:baseline}. We attribute this gap to the absence of the MPC's
constraint-aware optimization. The parameters predicted by VIGOR-MPC are used
by a controller that explicitly enforces the dynamics together with the track and
obstacle boundaries, so that even an imperfect parameter prediction still yields a
feasible and safe trajectory. The end-to-end policy, by contrast, must encode this
behavior implicitly in its weights and has no mechanism to enforce constraint
satisfaction at deployment.

{\textbf{Effect of Teacher Horizon Length $N_L$ on Lap Time and Per-step Solve Time:} As shown in Table~\ref{tab:teacher-horizon}, the teacher's lap time mproves with increasing horizon length, saturating at $N_L\!\approx\!30$, while solve time keeps increasing; shortening the horizon
reduces solve time but degrades lap time. Because of this solve time, the teacher serves purely as an
offline oracle that generates imitation targets during training and is never deployed on hardware. Since perception and control share a computational budget, cutting control cost by more than $90\%$ frees resources for perception.

\begin {figure}[H]
\centering
\begin{tikzpicture}[scale=0.85]
    \node (pic1) at (-2.4, 0.0) {\includegraphics[width=.29\textwidth]{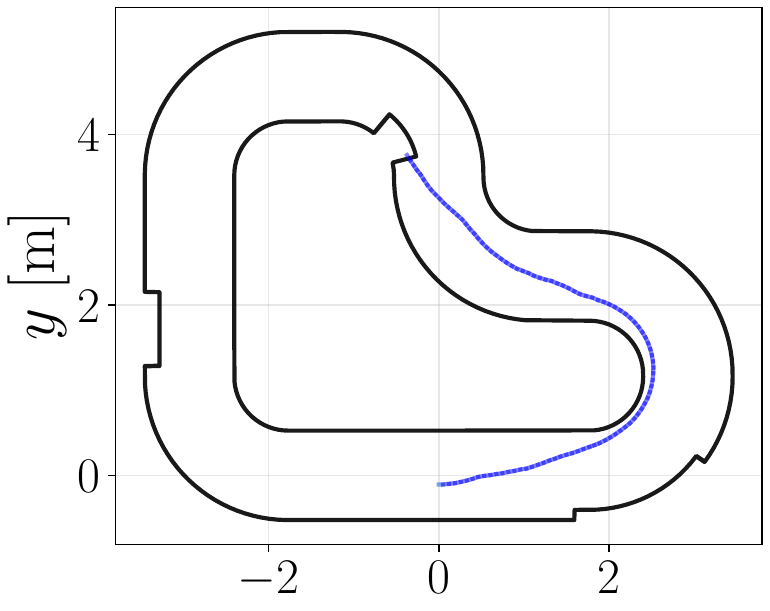}};
    \node (pic2) at (2.25, 0.0) {\includegraphics[width=.26\textwidth]{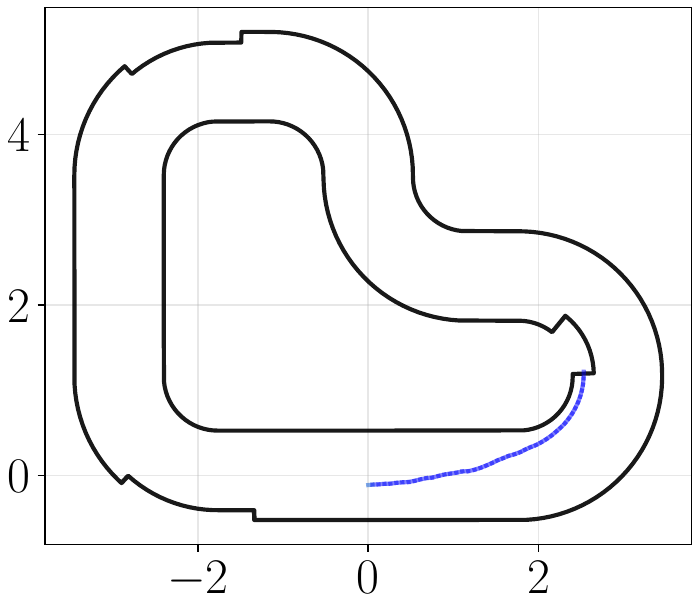}};
    \node (pic3) at (6.8, 0.0) {\includegraphics[width=.265\textwidth]{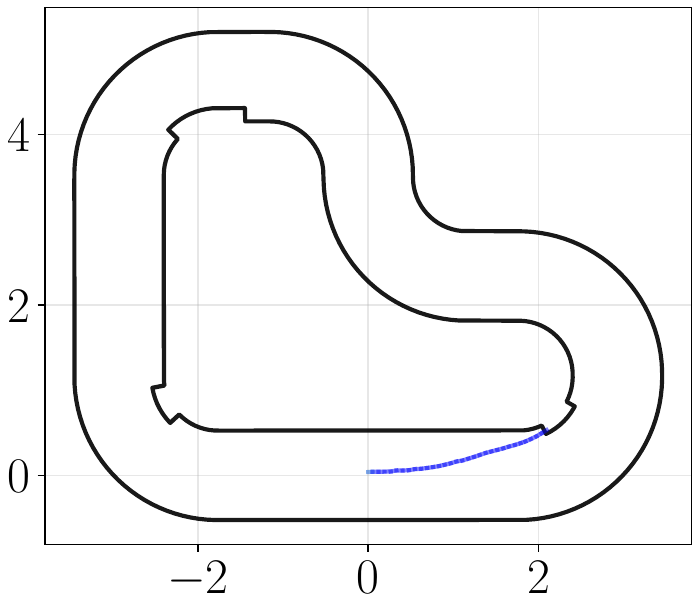}};
    \node (pic4) at (-2.4, -3.8) {\includegraphics[width=.29\textwidth]{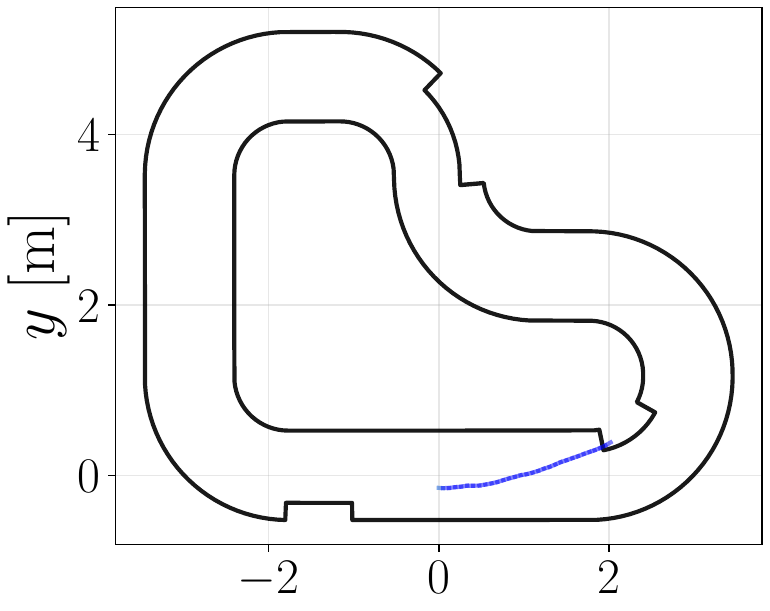}};
    \node (pic5) at (2.25, -3.8) {\includegraphics[width=.265\textwidth]{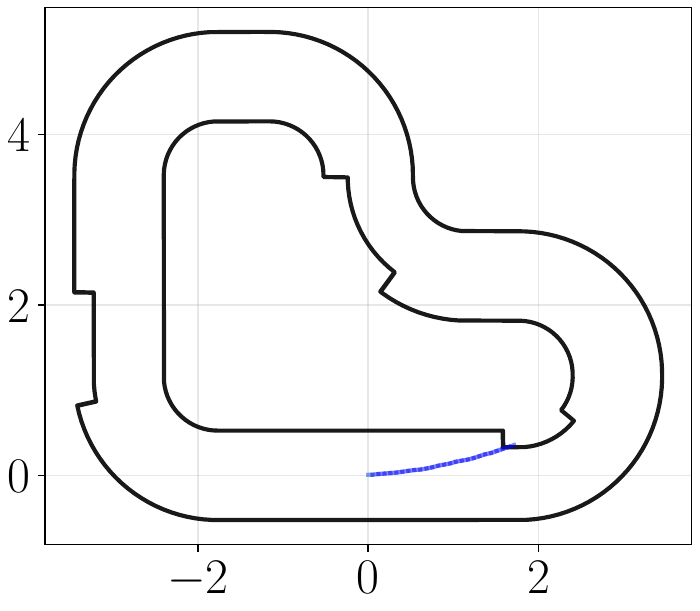}};
    \node (pic6) at (6.8, -3.8) {\includegraphics[width=.265\textwidth]{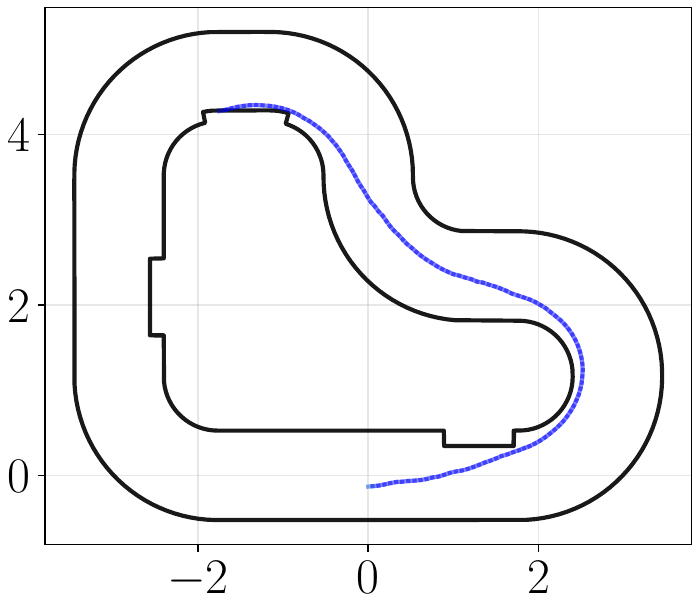}};
    \node (pic7) at (-2.4, -7.8) {\includegraphics[width=.29\textwidth]{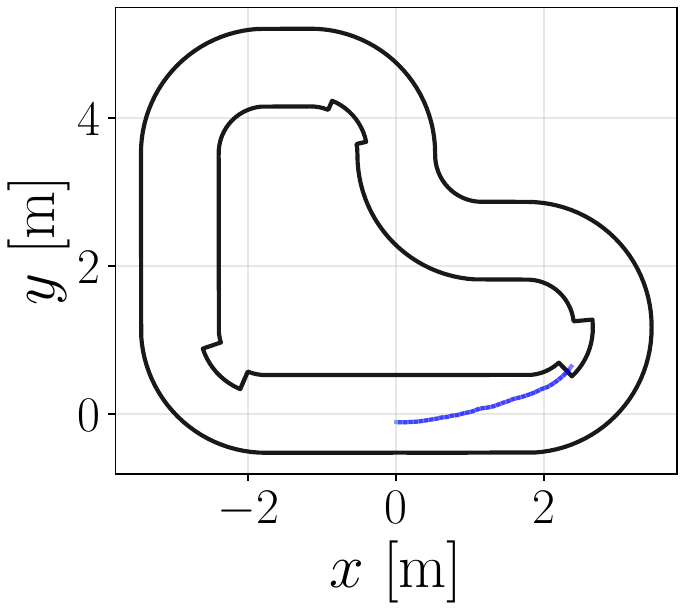}};
    \node (pic8) at (2.25, -7.8) {\includegraphics[width=.26\textwidth]{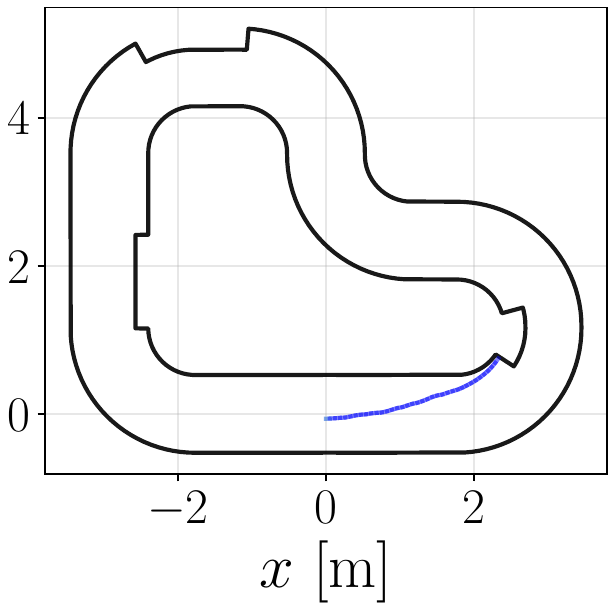}};
    \node (pic9) at (6.8, -7.8) {\includegraphics[width=.26\textwidth]{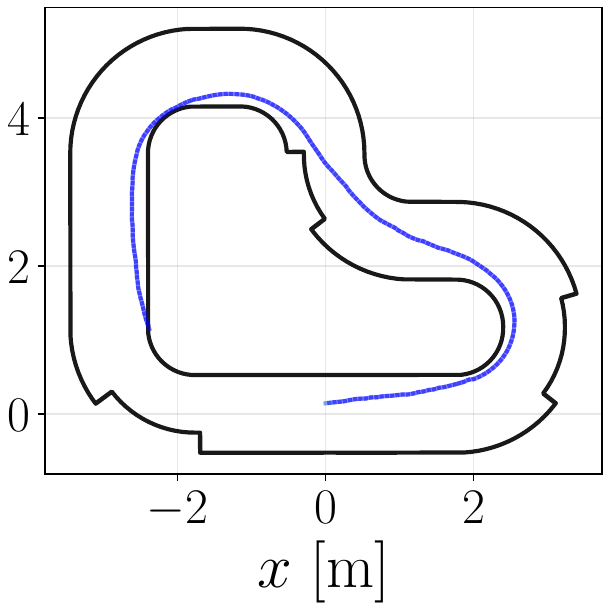}};

 \node (pic2) at (2.2, 3.3) {\includegraphics[width=.4\textwidth]{colorbar.pdf}};

\end{tikzpicture}
    \caption{{Closed-loop employment of the end-to-end baseline on different obstacle layouts (with randomly placed obstacle configurations corresponding to those shown in Figure~\ref{fig:addtnl_results}).}} 
    \vspace{-1em}
\label{fig:baseline}
\end{figure} 

\begin{table}[t]
\centering
\caption{{Lap-time and
per-step solve time of long horizon MT-MPC with increasing  horizon length,
indicating the used horizon length with a cross.
} }
\label{tab:teacher-horizon}
\begin{tabular}{lcccc}
\toprule
$N_L$          & $15$   & $25^{\dagger}$ & $30$    & $40$    \\
\midrule
Lap time [s]   & $7.46$ & $7.36$         & $7.33$  & $7.33$  \\
Solve time [ms] & $537.3$ & $978.2$       & $1284.5$ & $1982.3$ \\
\bottomrule
\end{tabular}
\end{table}}

\subsection{Additional Real-World Results}
\label{subsec:add_hard_results} In the following, we provide further experimental results obtained with our real-world F1-Tenth car, including, further obstacle configurations, comparisons to the short horizon MPC baseline $MPC_{N_S}$, and generalization capabilities. To not clutter our visualization, only one lap per obstacle configuration is provided. To demonstrate how our approach successfully handles dynamically changing obstacle configurations, we refer to our supplementary material and the contained video. 

\begin{figure}[htb!]
    \centering
    \includegraphics[width=0.95\linewidth]{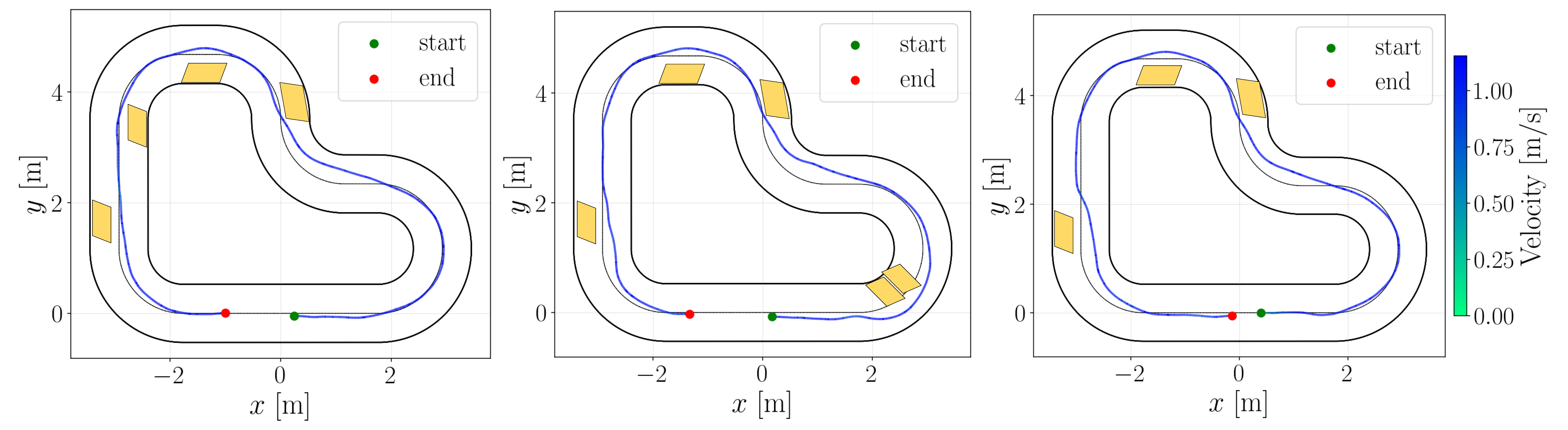}
    \vspace{-0.5em}
    \caption{Visualization of additional successfully handled obstacle configurations of VIGOR-MPC in closed-loop deployment on our real-world setup.}
    \vspace{-0.5em}\label{fig:add_results_hardware_obstacels}
\end{figure}

\textbf{Further Obstacle Configurations:}
In Figure~\ref{fig:add_results_hardware_obstacels} we illustrate a selection of successfully handled obstacle configurations, applying VIGOR-MPC in closed loop on our hardware setup. Note that all of these configuration results were collected in a continuous run in which we removed and replaced some of the obstacles.

\begin{figure}[htb!]
    \centering
    \includegraphics[width=0.95\linewidth]{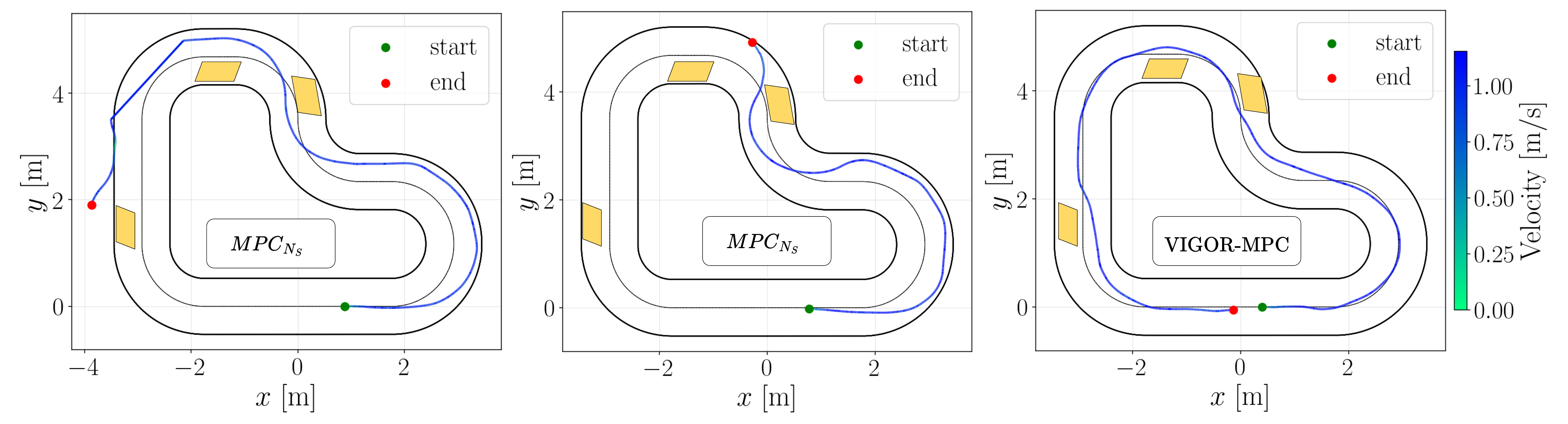}
    \vspace{-0.5em}
    \caption{Comparison of our short horizon MPC baseline $MPC_{N_S}$ and VIGOR-MPC in closed-loop deployment on our real-world setup. The $MPC_{N_S}$ laps are visualized in the image on the left and the one in the middle, VIGOR-MPC is deployed in the image on the right}
    \vspace{-0.5em}\label{fig:add_results_hardware_compare}
\end{figure}

\textbf{Comparison with Short Horizon MPC:} A comparision of our proposed method with the short horizon baseline MPC, $MPC_{N_S}$, both employing the same centerline estimation and obstacle detection pipeline, is provided in Figure~\ref{fig:add_results_hardware_compare}. Leveraging a comparable amount of compute, VIGOR-MPC successfully manages to complete the lap, while the short horizon baseline MPC, fails and crashes into the track boundary at various instances. 

\begin{figure}[htb!]
    \centering
    \includegraphics[width=0.85\linewidth]{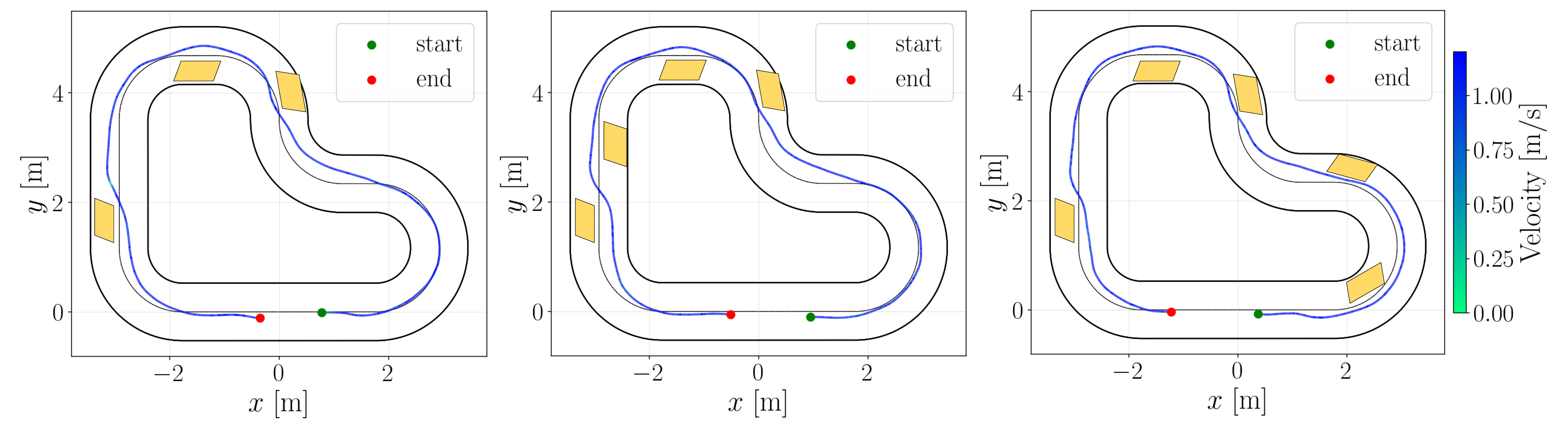}
    \vspace{-0.0em}
    \caption{Deployment of VIGOR-MPC on our hardware track, with a model that has only been trained on the leftmost track of Figure~\ref{fig:generalization} and dynamical obstacle replacement.}
    \vspace{0.5em}\label{fig:add_results_hardware_generalize}
\end{figure}

\textbf{Generalization Capabilities:}
Finally, we also demonstrate the generalization capabilities of VIGOR-MPC on hardware. Therefore, we executed a model, solely trained on the leftmost track in Figure~\ref{fig:generalization}, on our hardware track. Once again, we replaced the obstacles dynamically during execution. Three laps with different obstacle configurations are visualized in Figure~\ref{fig:add_results_hardware_generalize}.

\newpage

\subsection{Illustration of Training Behavior}

\begin {figure}[H]
\centering
\begin{tikzpicture}[scale=0.7]

    \node (pic1) at (-3, 0.0) {\includegraphics[width=.28\textwidth]{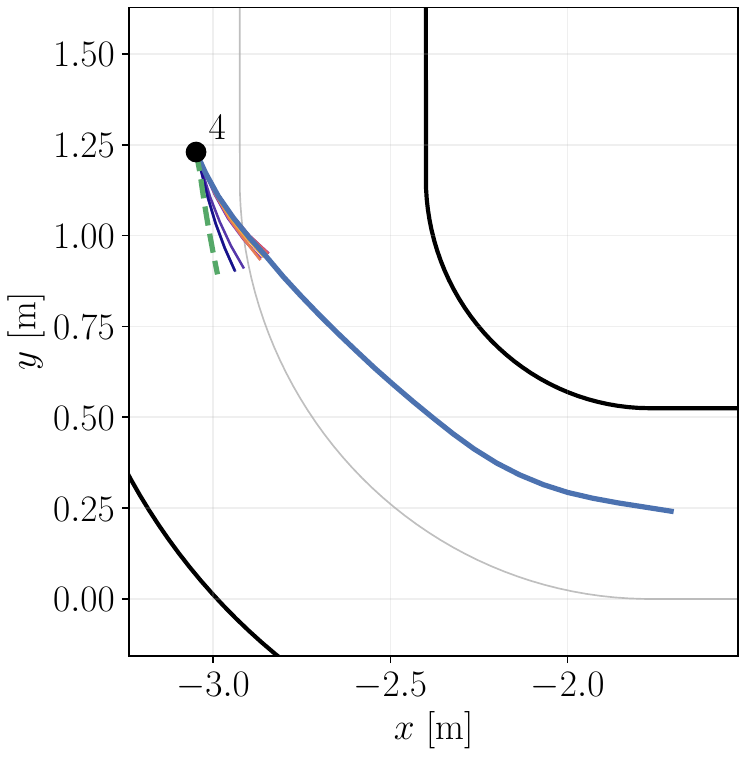}};

    \node (pic2) at (4, 0.0) {\includegraphics[width=.28\textwidth]{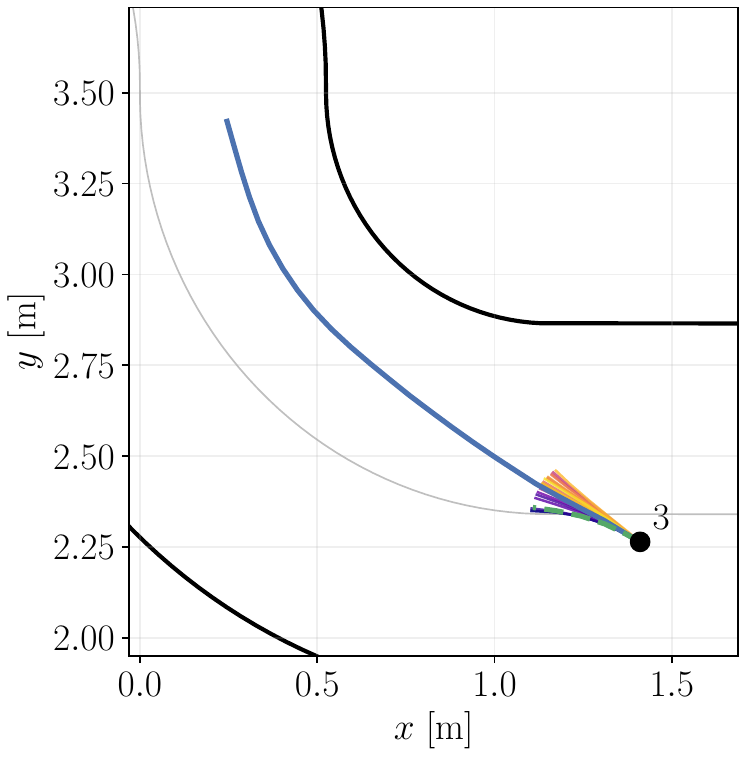}};

    \node (pic3) at (-3, 6.0) {\includegraphics[width=.28\textwidth]{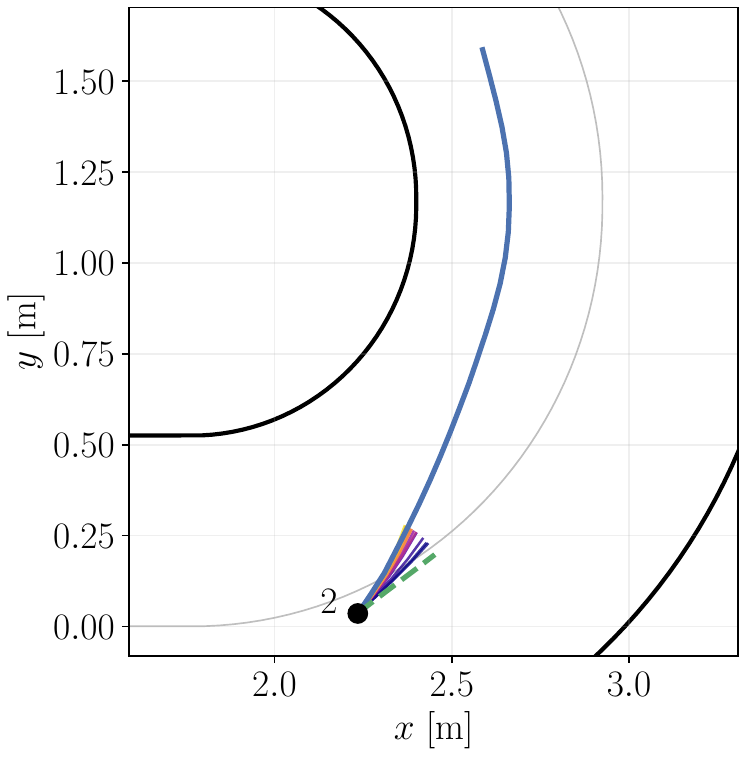}};

    \node (pic4) at (4, 6.0) {\includegraphics[width=.28\textwidth]{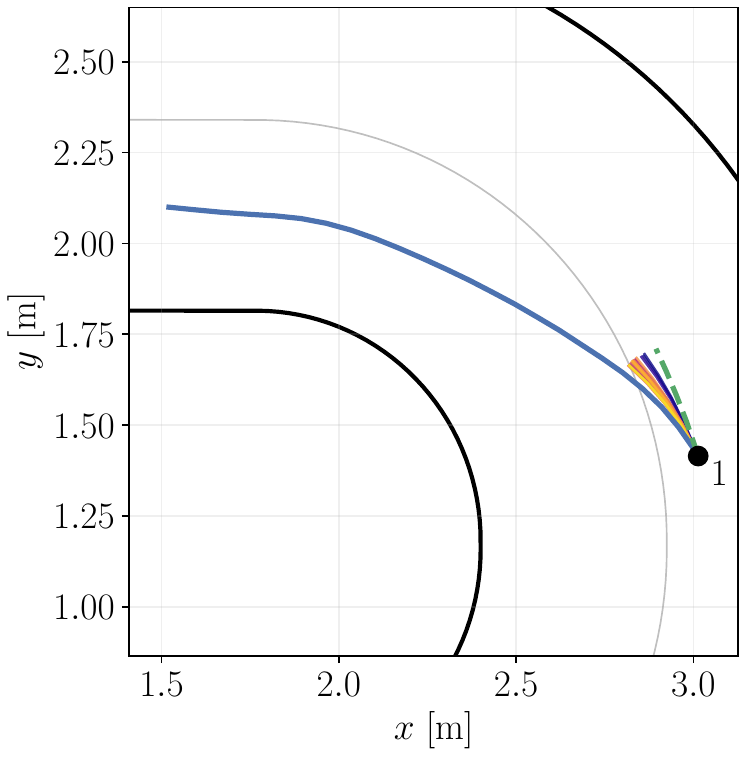}};

 \node (pic5) at (1.0, 14.8) {\includegraphics[width=.7\textwidth]{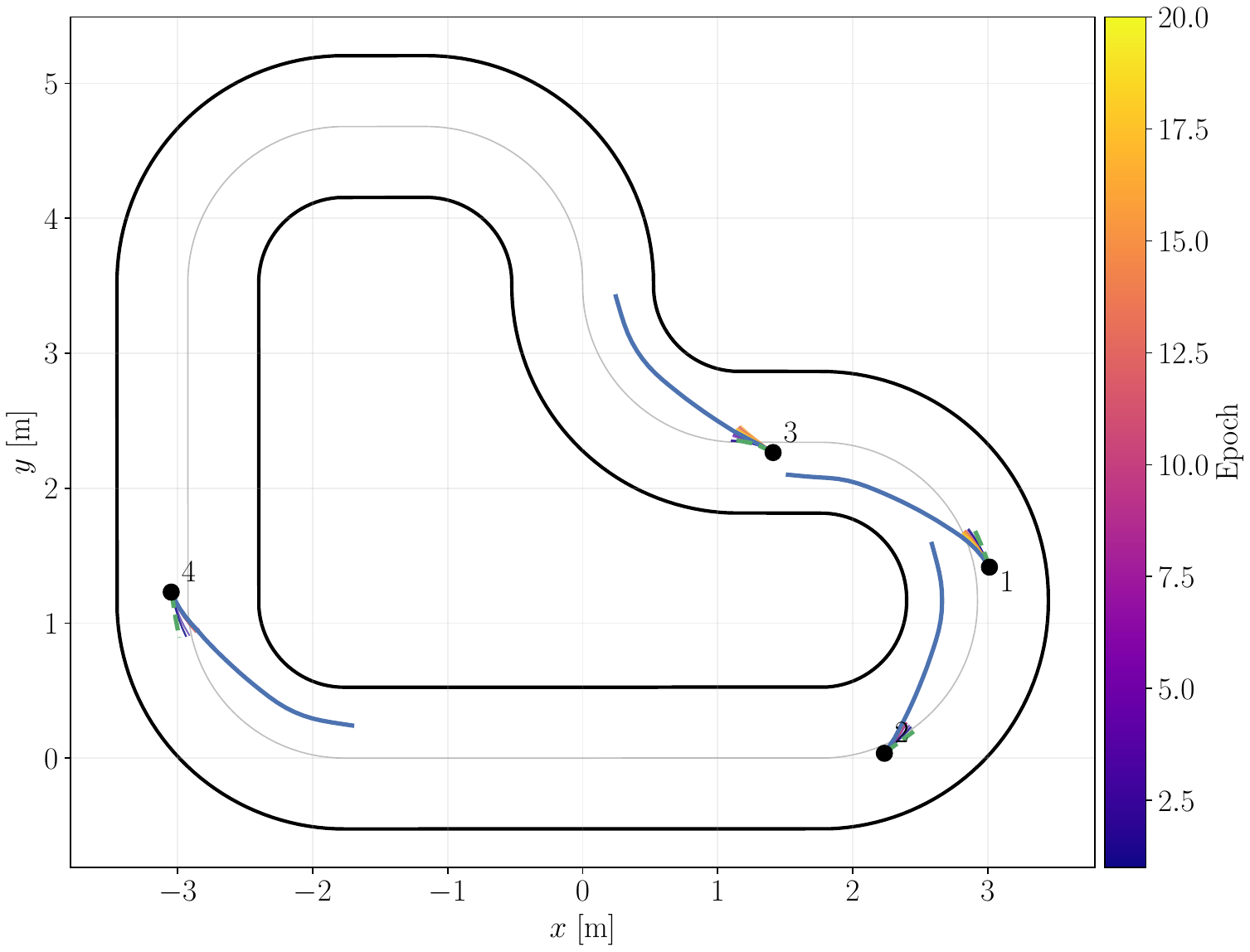}};

\end{tikzpicture}
    \caption{{Open-loop VIGOR-MPC prediction as training progresses against long horizon MT-MPC open-loop prediction (solid, in blue) and short-horizon MPC open-loop prediction (dashed, in green).}} 
\label{fig:zooms}
\end{figure}

\begin{figure}[H]
    \centering
    \includegraphics[width=0.68\linewidth]{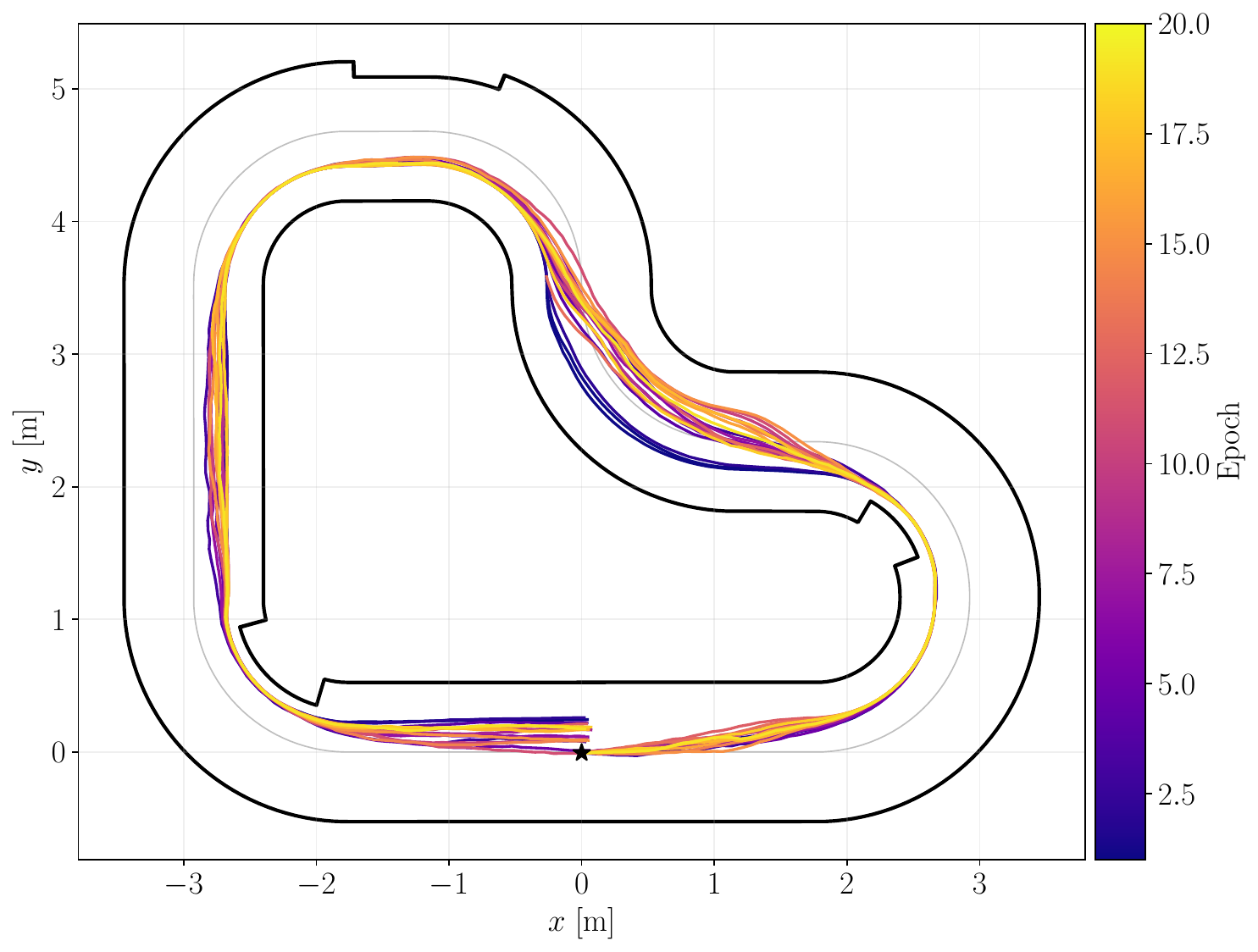}
    \caption{Evolution of the VIGOR-MPC trajectories across epochs.}
    \label{fig:traj_evol}
\end{figure}

We show how the trajectory of the car changes as the training progresses in Figure~\ref{fig:traj_evol} and how the open loop prediction changes with respect to the perfect teacher and short horizon MPC in Figure~\ref{fig:zooms}. The figures demonstrate that, as training progresses, the behavior of VIGOR-MPC shifts from short-horizon MPC to long-horizon MT-MPC behavior.

\label{subsec:add_train_results}

\subsection{Extension to Quadrotor Dynamics}
\label{apx:drone_results}

{To demonstrate that VIGOR-MPC is not specific to car racing, we extended our simulation pipeline to account for quadrotor dynamics. More specifically, we adapted the dynamics of \cite{kumar2017path}, which describe the motion of the quadrotor relative to a reference path in a Frenet-frame, and simplified them as follows. First, we neglect the motor and attitude dynamics, assuming that the inner attitude loop is sufficiently fast. The control inputs are therefore the desired accelerations $a_T$, $a_N$, $a_B$ along the tangential, normal, and binormal directions of the path, together with the yaw rate $\omega_\psi$, so that each translational axis evolves as a double integrator. Second, we set the path torsion to zero, $\tau = 0$. This decouples the lateral offset $d$ from the out-of-plane offset $n$, so that $n$ describes the altitude of the quadrotor relative to the track plane. Third, the heading $\psi$ is decoupled from the translational motion and simply integrates the commanded yaw rate. With state $x = [s, d, n, v_T, v_N, v_B, \psi]^\top$ and input $u = [a_T, a_N, a_B, \omega_\psi]^\top$, discretizing with the forward Euler method and sampling time $\Delta t$ gives
\begin{equation}
\label{eq:quad_dynamics}
\begin{aligned}
    s(k+1)        &= s(k) + T \, \frac{v_{T}(k)}{1 - \kappa(s(k))\, d(k)}, \\
    d(k+1)        &= d(k) + T \, v_{N}(k), \\
    n(k+1)        &= n(k) + T \, v_{B}(k), \\
    v_{T}(k+1)      &= v_{T}(k) + T \, a_{T}(k), \\
    v_{N}(k+1)      &= v_{N}(k) + T \, a_{N}(k), \\
    v_{B}(k+1)      &= v_{B}(k) + T \, a_{B}(k), \\
    \psi(k+1)     &= \psi(k) + T \, \omega_{\psi}(k).
\end{aligned}
\end{equation}
As in the car racing setting, the state is augmented with soft-constraint penalty terms. These keep the lateral offset within the track boundaries, $|d| \leq \tfrac{1}{2} w$, and the tangential velocity within $0 \leq v_T \leq v_{\max}$. The inputs are bounded by $|a_i| \leq a_{\max}$ for $i \in \{T, N, B\}$ and $|\omega_\psi| \leq \omega_{\psi,\max}$, where $a_{\max} = 8 \mathrm{m}/\mathrm{s}^2$ and $\omega_{\psi,\max} = 3 \mathrm{rad}/\mathrm{s}$. With this setup, we performed the following two experiments.} 

\begin{figure}[H]
    \centering
    \includegraphics[width=0.98\linewidth]{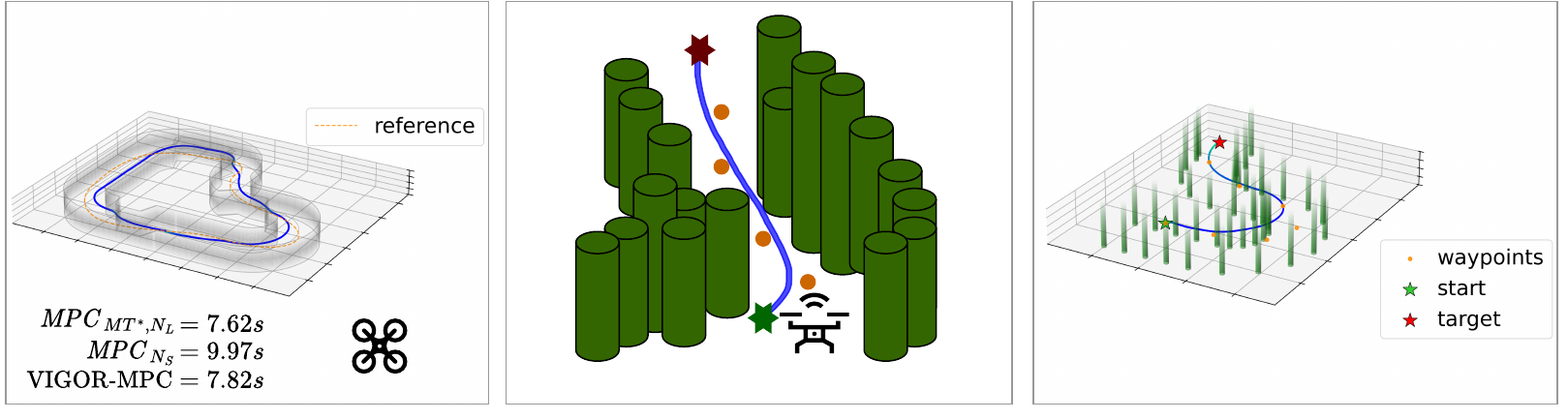}
    \caption{Visualization of the quadrotor experiments. Left image: quadrotor racing application in 3D, with an illustration of VIGOR-MPC's racing line and the lap times of $MPC_{MT^*,N_L}$, $MPC_{N_S}$, and VIGOR-MPC shown at the bottom, where $N_L = 25$ and $N_S = 6$. Middle image: explanatory figure illustrating how a reference-tracking task in a constrained environment can be reformulated as a path-following task. Right image: application of VIGOR-MPC with quadrotor dynamics to solve a reference-tracking task in a pillared environment.}
    \label{fig:drone_experiments}
\end{figure}

\textbf{Racing in 3D:} {We extend the racing track to 3D and augment the reference line with an altitude profile given by
$
n_{\mathrm{ref}}(s) = A \sin \left(\frac{2\pi}{\lambda}s \right)
$.
We choose an amplitude of $A = 0.15,\mathrm{m}$ and a wavelength of $\lambda = 6.0,\mathrm{m}$.
Using the long-horizon multi-trajectory MPC, $MPC_{MT^*,N_L}$, with a horizon length of $N_L = 25$ achieves a lap time of $7.62 \mathrm{s}$. However, running our VIGOR-MPC training pipeline, as described for the car example, VIGOR-MPC nearly matches this performance, finishing the lap within $7.82 \mathrm{s}$ with a horizon length of only $6$. In contrast, the baseline short-horizon MPC, $MPC_{N_S}$, using the same horizon length of $6$, requires $9.97 \mathrm{s}$ to complete the lap, resulting in a performance gap of more than $2 \mathrm{s}$. A visualization of the 3D racing track is provided in the leftmost image of Figure~\ref{fig:drone_experiments}.}

\textbf{Reference Tracking in 3D:} { Leveraging the fact that reference-tracking problems can be formulated as path-following problems, we demonstrate that VIGOR-MPC can also be applied to this task by connecting the provided reference points with trajectory segments. We consider a quadrotor operating in a pillared environment, where reference points are placed midway between consecutive pillars, and the quadrotor is required to come to a stop at the final reference point. A visualization of the tracking task and its execution in simulation is shown in the middle and rightmost images of Figure~\ref{fig:drone_experiments}.}

\end{document}